\documentclass[10pt,twocolumn,letterpaper]{article}

\usepackage{iccv}
\usepackage{times}
\usepackage{epsfig}
\usepackage{graphicx}
\usepackage{amsmath}
\usepackage{amssymb}

\usepackage{comment}
\usepackage{color}

\usepackage{multirow}
\usepackage{bbold}
\usepackage{bbm}

\makeatletter
\@namedef{ver@everyshi.sty}{}
\makeatother
\usepackage{tikz}

\usepackage{pgfplots}
\usetikzlibrary{spy,calc}
\usetikzlibrary{arrows.meta}
\usetikzlibrary{trees} 
\usetikzlibrary{shapes,arrows,decorations.markings}
\usetikzlibrary{intersections}
\usetikzlibrary{matrix}
\usetikzlibrary{datavisualization}

\usepackage{ctable}
\usepackage{booktabs}
\input{preamble}

\usepackage[export]{adjustbox}

\usepackage[pagebackref=true,breaklinks=true,colorlinks,bookmarks=false]{hyperref}

\iccvfinalcopy 

\def\iccvPaperID{8895} 

\ificcvfinal\fi

\begin{document}

\title{FMODetect: Robust Detection of Fast Moving Objects}

\author{Denys Rozumnyi$^{1,4}$\\
\and
Ji\v{r}\'\i{} Matas$^{4}$\\
\and
Filip \v{S}roubek$^{2}$\\
\and
Marc Pollefeys$^{1,3}$\\
\and \hspace{-1em}
Martin R. Oswald$^{1}$\\
\and
{\normalsize $^{1}$Department of Computer Science, ETH Zurich }
\and
{\normalsize  $^{3}$Microsoft Mixed Reality and AI Zurich Lab }
\and
{\normalsize $^{2}$UTIA, Czech Academy of Sciences}
\and
{\normalsize $^{4}$Visual Recognition Group, Czech Technical University in Prague}
}

\maketitle
\ificcvfinal\thispagestyle{empty}\fi

\begin{abstract}
We propose the first learning-based approach for fast moving objects detection. Such objects are highly blurred and move over large distances within one video frame. Fast moving objects are associated with a deblurring and matting problem, also called deblatting. We show that the separation of deblatting into consecutive matting and deblurring allows achieving real-time performance, i.e.~an order of magnitude speed-up, and thus enabling new classes of application. The proposed method detects fast moving objects as a truncated distance function to the trajectory by learning from synthetic data. For the sharp appearance estimation and accurate trajectory estimation, we propose a matting and fitting network that estimates the blurred appearance without background, followed by an energy minimization based deblurring. The state-of-the-art methods are outperformed in terms of recall, precision, trajectory estimation, and sharp appearance reconstruction. Compared to other methods, such as deblatting, the inference is of several orders of magnitude faster and allows applications such as real-time fast moving object detection and retrieval in large video collections.
\end{abstract}
 
\section{Introduction}
Fast moving objects (FMOs) are objects that look significantly blurred in images. 
Within the aperture of a single frame, they move at high speed over distances larger than their size.
FMOs typically appear as blurred streaks formed as a composition of the background and their blurred appearance. 
Other scenarios include fast rotations of objects like ventilators or propellers.
FMOs are common in everyday scenarios like fast cars, falling objects, flying insects, rain, and hailstorm.
They are mainly present in sports videos such as tennis, football, badminton, or other games with moving objects. 
Applications of FMO detection methods are temporal super-resolution, speed measurement, FMO retrieval, tracking, removal, and highlighting.

\begin{figure}[t]
\centering
\begin{tabular}{@{}c@{}}

\resizebox {0.47\textwidth}{!} {
\begin{tikzpicture}[zoomboxarray,
    zoomboxes right,
    zoomboxarray columns=1,
    zoomboxarray rows=1,
    connect zoomboxes,
    zoombox paths/.append style={thick, red}]
    \node [image node] { \includegraphics[width=0.25\textwidth]{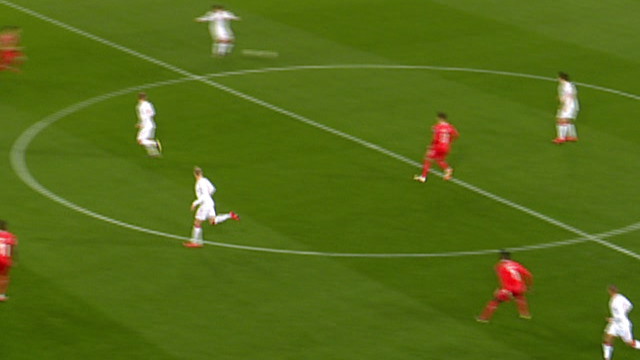} };
    \zoombox[magnification=10,color code=green]{0.405,0.85}
\end{tikzpicture}
}

\\

\resizebox {0.47\textwidth}{!} {
\begin{tikzpicture}[zoomboxarray,
    zoomboxes right,
    zoomboxarray columns=2,
    zoomboxarray rows=1,
    connect zoomboxes,
    zoombox paths/.append style={thick, red}]
    \node [image node] { \includegraphics[width=0.25\textwidth]{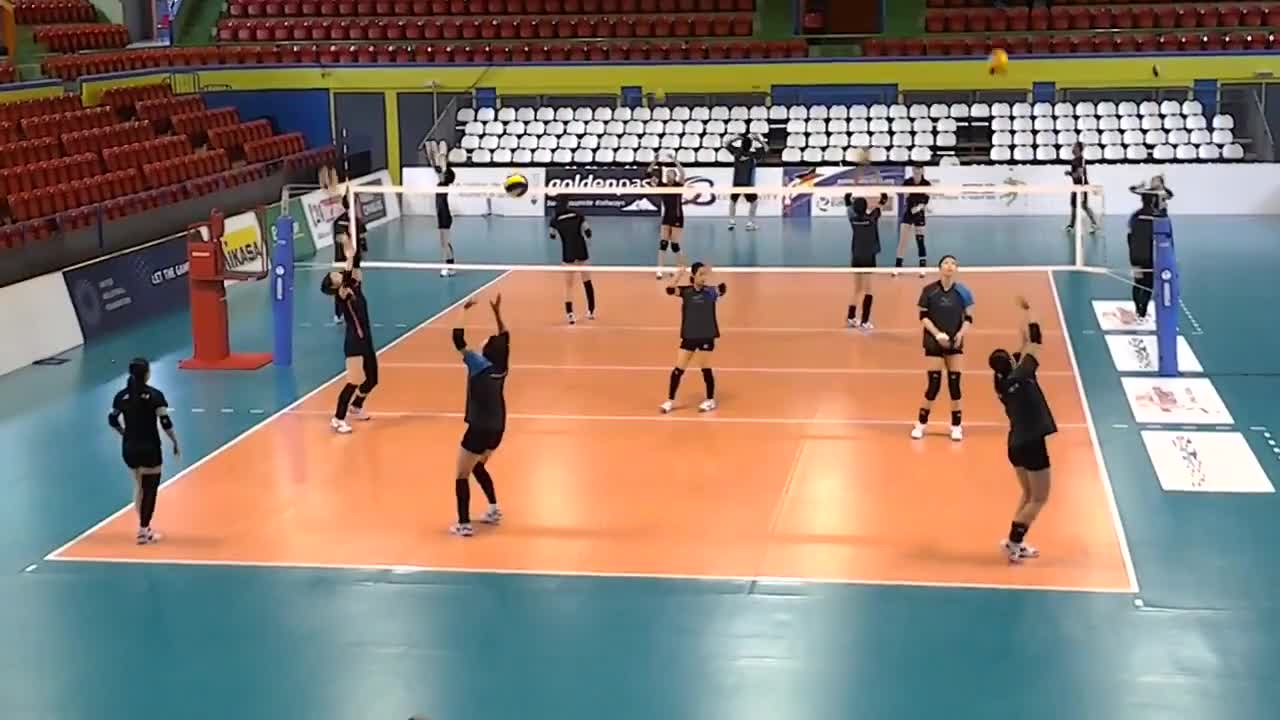} };
    \zoombox[magnification=10,color code=green]{0.67,0.77}
    \zoombox[magnification=10,color code=green]{0.78,0.91}
\end{tikzpicture}
}

\\ 


\end{tabular}

\caption{Examples of automatically detected fast moving objects (FMOs) by the proposed method on soccer (top) and volleyball (bottom) videos. 
}
\label{fig:intro}
\end{figure}

\newcommand{\mframe}[3]{%
\begin{tikzpicture}[zoomboxarray,
    zoomboxes below,
    black and white,
    zoomboxarray columns=1,
    zoomboxarray rows=1,
    connect zoomboxes,
    zoombox paths/.append style={thick, red}]
    \node [image node] { \includegraphics[width=0.19\textwidth]{#1} };
    \zoombox[magnification=#2]{#3}
\end{tikzpicture}
}

\begin{figure*}[t]
\centering
\scriptsize
\begin{tabular}{@{}c@{}c@{}c@{}c@{}c@{}}
\mframe{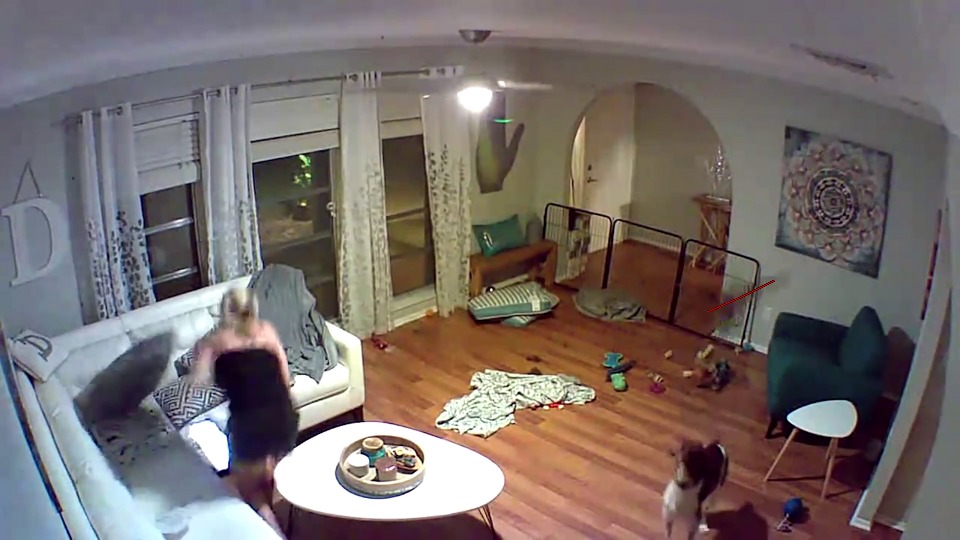}{5}{0.75,0.45} &
\mframe{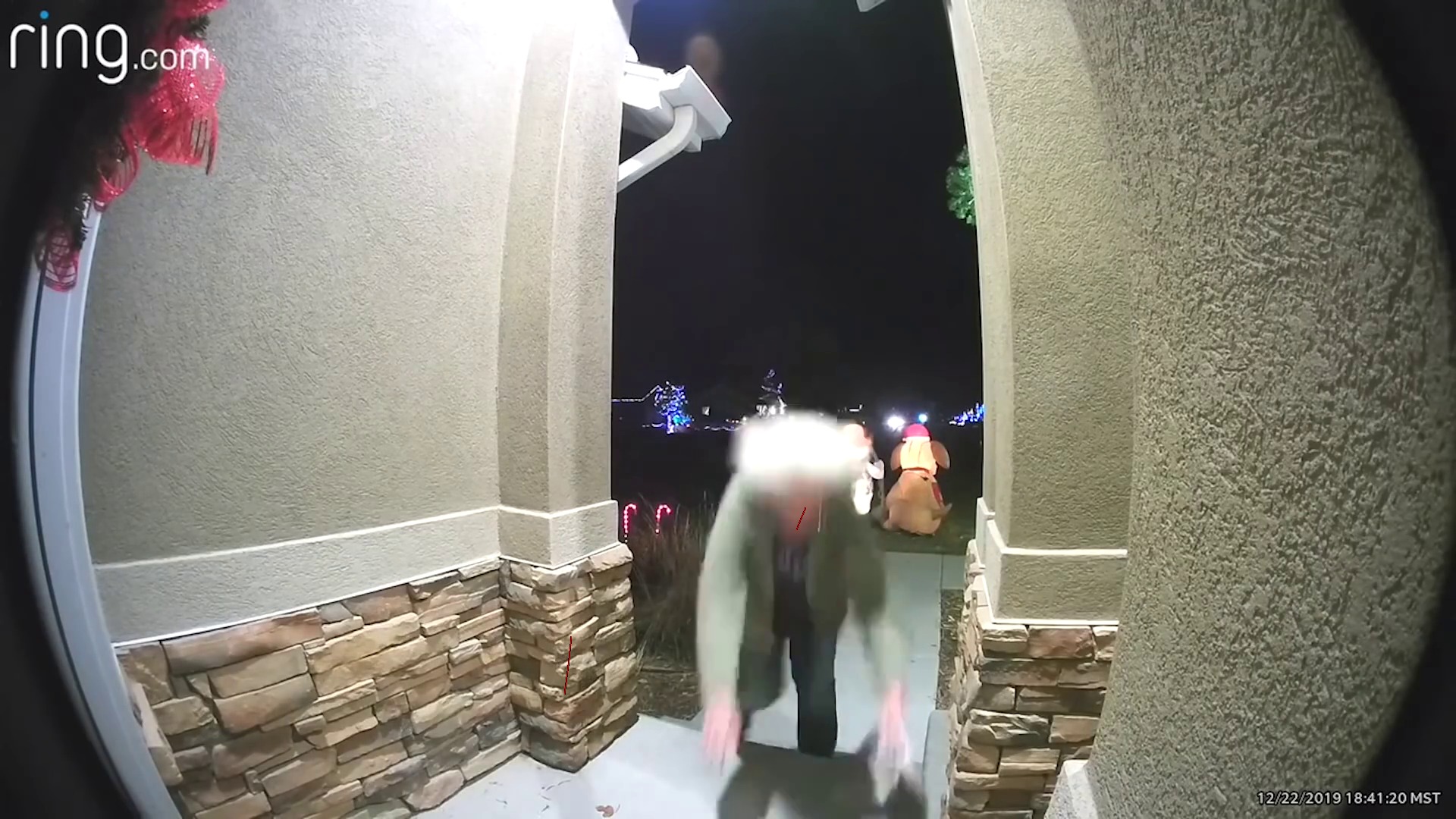}{9}{0.55,0.38} &
\mframe{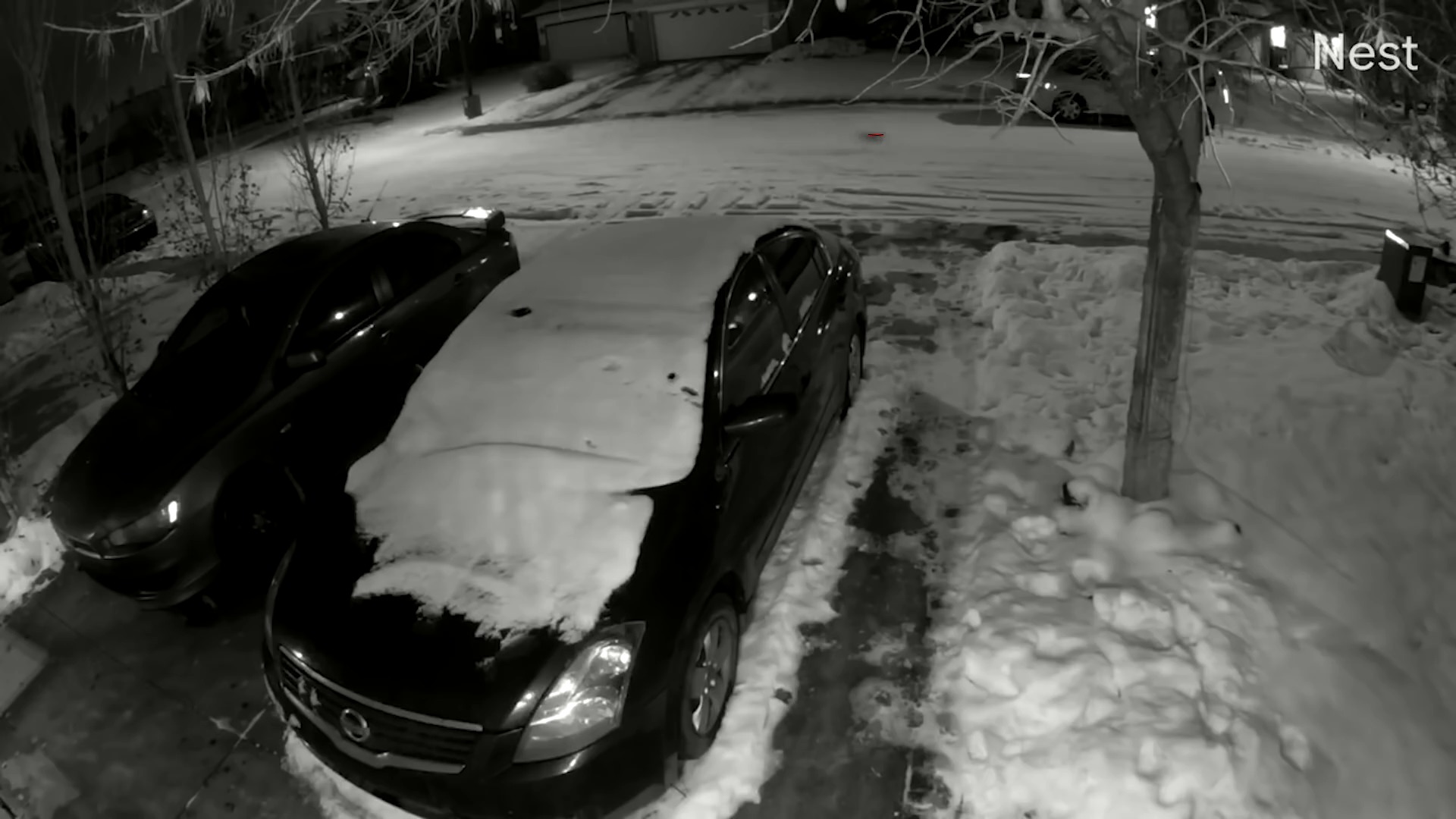}{11}{0.6,0.83} &
\mframe{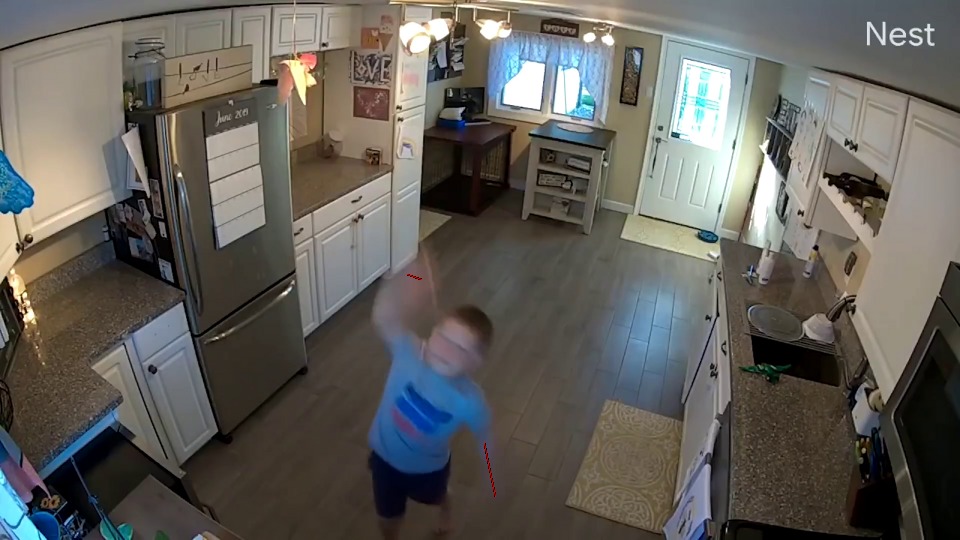}{5}{0.45,0.5} &
\mframe{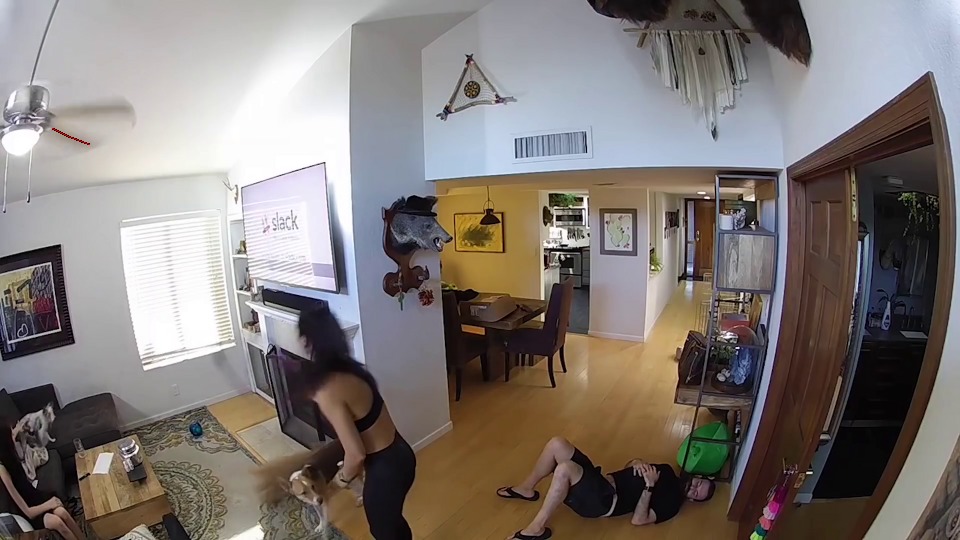}{5}{0.1,0.75}   \\

\vspace{3ex}
Bird [\href{https://youtu.be/CkVJyAKwByw?t=9}{0:09}]  & 
Head [\href{https://youtu.be/CkVJyAKwByw?t=87}{1:27}] &
Rabbit [\href{https://youtu.be/CkVJyAKwByw?t=160}{2:40}] &
Hand [\href{https://youtu.be/CkVJyAKwByw?t=232}{3:52}] & 
Ceiling fan [\href{https://youtu.be/CkVJyAKwByw?t=303}{5:03}]  \\

\mframe{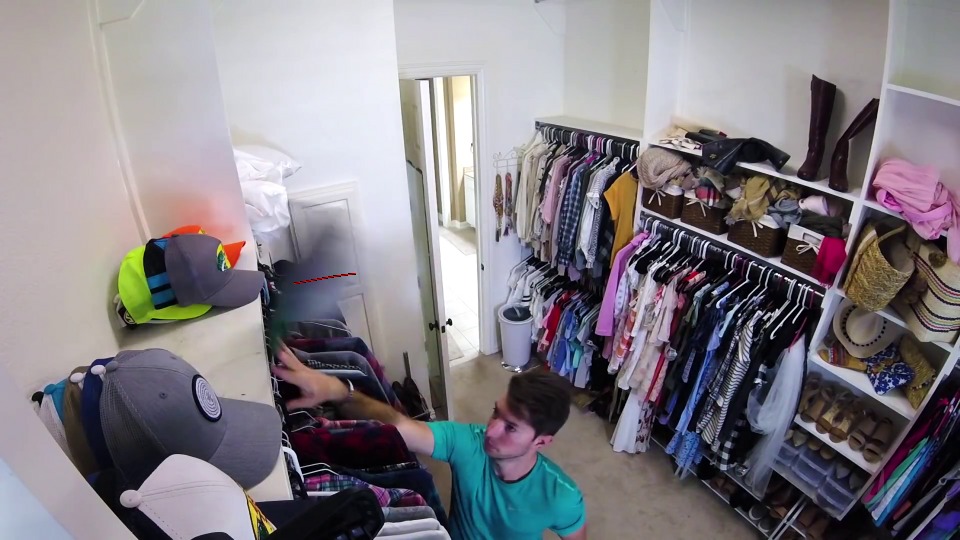}{3}{0.35,0.5} &
\mframe{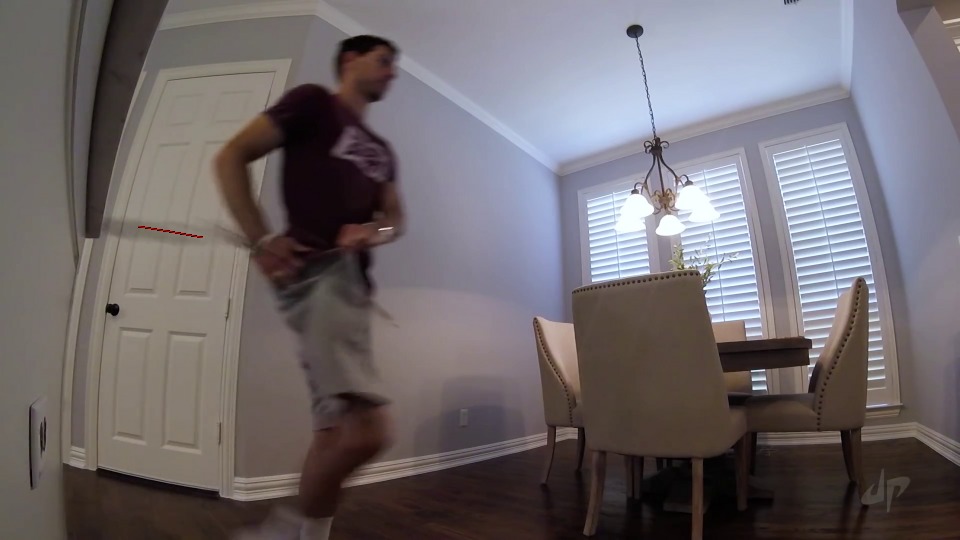}{4}{0.2,0.55} &
\mframe{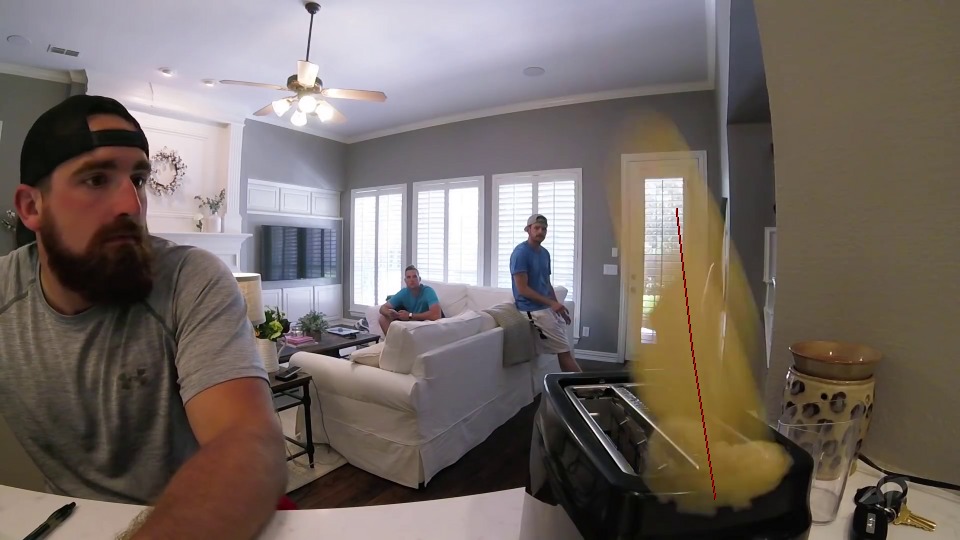}{1.5}{0.65,0.35} &
\mframe{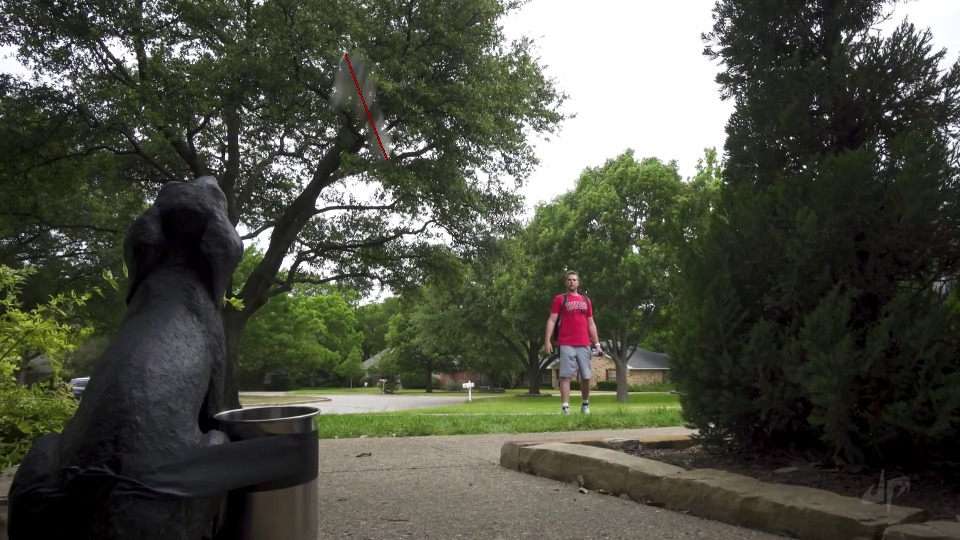}{4}{0.4,0.8} &
\mframe{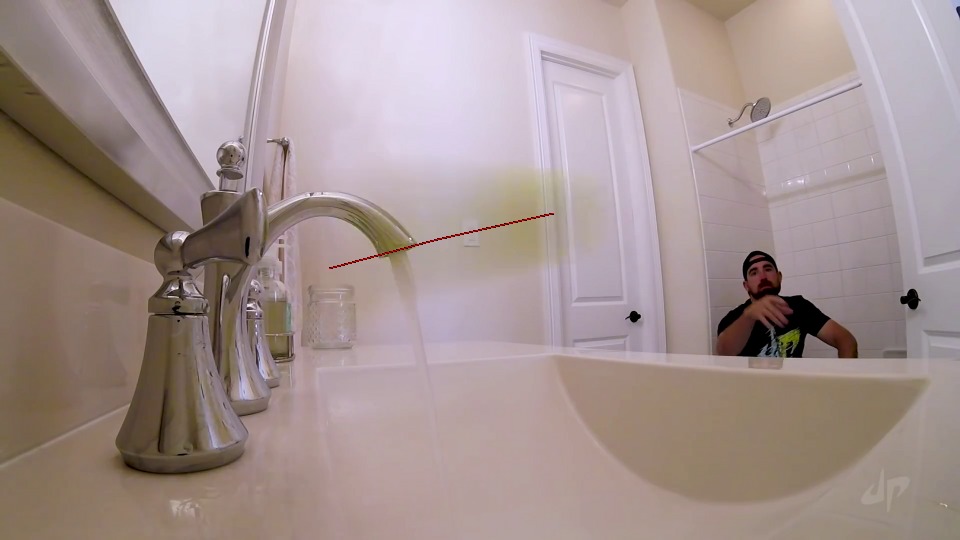}{2}{0.45,0.55} \\

Cap [\href{https://youtu.be/A2FsgKoGD04?t=61}{1:01}]  & Keys [\href{https://youtu.be/A2FsgKoGD04?t=109}{1:49}] & Toast [\href{https://youtu.be/A2FsgKoGD04?t=214}{3:34}] & Newspaper [\href{https://youtu.be/A2FsgKoGD04?t=219}{3:39}] & Tennis ball [\href{https://youtu.be/A2FsgKoGD04?t=245}{4:05}] \\

\end{tabular}

\caption{Examples of fast moving object (FMO) retrieval on YouTube videos by our method running in real-time. \textbf{Top:} frames from a 10-minute surveillance footage [\href{https://youtu.be/CkVJyAKwByw}{youtu.be/CkVJyAKwByw}]. \textbf{Bottom:} frames from a 5-minute trick-shot video [\href{https://youtu.be/A2FsgKoGD04}{youtu.be/A2FsgKoGD04}]. For each image, the detected FMO and its trajectory is shown as a red line in a close-up below. The  object description contains a digitally clickable link to the timestamp when the event happened, in $[$minutes:seconds$]$ format. }
\label{fig:retrieval}
\end{figure*}


\begin{figure}[t]
\centering
\begin{tabular}{@{}c@{}c@{}}

\resizebox {0.22\textwidth}{!} {\begin{tikzpicture}
\begin{axis}[legend pos=north west,
    ymin=0, ymax=0.8,
    minor y tick num = 3,
    area style,
    ]
\addplot+[ybar interval,mark=no,blue!60,fill opacity=0.4,] plot coordinates { 
( 0.0 ,  0.0029846291598268917 )
( 0.1 ,  0.012236979555290255 )
( 0.2 ,  0.0123862110132816 )
( 0.30000000000000004 ,  0.019101626622892106 )
( 0.4 ,  0.030741680346216983 )
( 0.5 ,  0.04223250261155052 )
( 0.6000000000000001 ,  0.09237427249664229 )
( 0.7000000000000001 ,  0.15520071631099835 )
( 0.8 ,  0.31965378301746006 )
( 0.9 ,  0.3130875988658409 )
( 1.0 ,  0.3130875988658409 )
};

\addplot+[ybar interval,mark=no,red!60,fill opacity=0.4,] plot coordinates { 
( 0.0 ,  0.0 )
( 0.1 ,  0.0 )
( 0.2 ,  0.004258943781942078 )
( 0.30000000000000004 ,  0.004258943781942078 )
( 0.4 ,  0.0068143100511073255 )
( 0.5 ,  0.019591141396933562 )
( 0.6000000000000001 ,  0.03492333901192504 )
( 0.7000000000000001 ,  0.075809199318569 )
( 0.8 ,  0.21379897785349233 )
( 0.9 ,  0.6405451448040886 )
( 1.0 ,  0.6405451448040886 )
};

\addplot+[ybar interval,mark=no,cyan!60,fill opacity=0.4,] plot coordinates { 
(0, 0.15468)
(0.1, 0.33115)
(0.2, 0.20697)
(0.3, 0.11983)
(0.4, 0.058824)
(0.5, 0.076253)
(0.6, 0.026144)
(0.7, 0.017429)
(0.8, 0.0043573)
(0.9, 0)
(1, 0.0043573)
};

\legend{Got10k~\cite{got10k},LaSOT~\cite{lasot},TbD~\cite{tbd}}
\end{axis}
\end{tikzpicture}}
 
 &
 
 \resizebox {0.22\textwidth}{!} {\begin{tikzpicture}
\begin{axis}[legend style={at={(0.3,0.86)},anchor=west},
    ymin=0, ymax=0.8,
    minor y tick num = 3,
    area style,
    ]
\addplot+[ybar interval,mark=no,blue!60,fill opacity=0.4,] plot coordinates { 
( 0.0 ,  0.5719023213747362 )
( 0.1 ,  0.21525474826650587 )
( 0.2 ,  0.07491709375942117 )
( 0.30000000000000004 ,  0.042357552004823634 )
( 0.4 ,  0.020198974977389206 )
( 0.5 ,  0.013566475731082304 )
( 0.6000000000000001 ,  0.013415737111848056 )
( 0.7000000000000001 ,  0.010551703346397347 )
( 0.8 ,  0.005878806150135665 )
( 0.9 ,  0.00618028338860416 )
( 1.0 ,  0.005577328911667169 )
( 1.1 ,  0.0027132951462164605 )
( 1.2000000000000002 ,  0.00482363581549593 )
( 1.3 ,  0.0010551703346397348 )
( 1.4000000000000001 ,  0.0018088634308109737 )
( 1.5 ,  0.001658124811576726 )
( 1.6 ,  0.0027132951462164605 )
( 1.7000000000000002 ,  0.0018088634308109737 )
( 1.8 ,  0.0021103406692794696 )
( 1.9000000000000001 ,  0.0015073861923424782 )
};

\addplot+[ybar interval,mark=no,red!60,fill opacity=0.4,] plot coordinates { 
( 0.0 ,  0.7381370826010545 )
( 0.1 ,  0.14323374340949033 )
( 0.2 ,  0.05008787346221441 )
( 0.30000000000000004 ,  0.0210896309314587 )
( 0.4 ,  0.01845342706502636 )
( 0.5 ,  0.007029876977152899 )
( 0.6000000000000001 ,  0.006151142355008787 )
( 0.7000000000000001 ,  0.0035149384885764497 )
( 0.8 ,  0.0008787346221441124 )
( 0.9 ,  0.0026362038664323375 )
( 1.0 ,  0.0 )
( 1.1 ,  0.0017574692442882249 )
( 1.2000000000000002 ,  0.0017574692442882249 )
( 1.3 ,  0.0017574692442882249 )
( 1.4000000000000001 ,  0.0008787346221441124 )
( 1.5 ,  0.0008787346221441124 )
( 1.6 ,  0.0008787346221441124 )
( 1.7000000000000002 ,  0.0 )
( 1.8 ,  0.0 )
( 1.9000000000000001 ,  0.0008787346221441124 )
};

\addplot+[ybar interval,mark=no,cyan!60,fill opacity=0.4,] plot coordinates { 
(0, 0.0084926)
(0.1, 0.0042463)
(0.2, 0)
(0.3, 0)
(0.4, 0.010616)
(0.5, 0.014862)
(0.6, 0.012739)
(0.7, 0.012739)
(0.8, 0.03397)
(0.9, 0.031847)
(1, 0.036093)
(1.1, 0.016985)
(1.2, 0.010616)
(1.3, 0.012739)
(1.4, 0.021231)
(1.5, 0.012739)
(1.6, 0.014862)
(1.7, 0.03397)
(1.8, 0.021231)
(1.9, 0.66879)
(2, 0.66879)
};

\legend{Got10k~\cite{got10k},LaSOT~\cite{lasot},TbD~\cite{tbd}}
\end{axis}
\end{tikzpicture}}

\\

(a) IoU($\text{GT}_{t}$,$\text{GT}_{t+1}$) & (b) Normalized speed \\[0.5em] 

\end{tabular}

\caption{Histograms of (a) Intersection over Union (IoU) scores between consecutive ground truth bounding boxes and (b) speed normalized by the object size for datasets GOT-10k~\cite{got10k}, LaSOT~\cite{lasot}, and TbD~\cite{tbd} that focuses on FMOs. FMOs are present in GOT-10k and LaSOT datasets, but never annotated,~\ie trackers are not tested on fast moving objects.}
\label{fig:hist}
\end{figure}
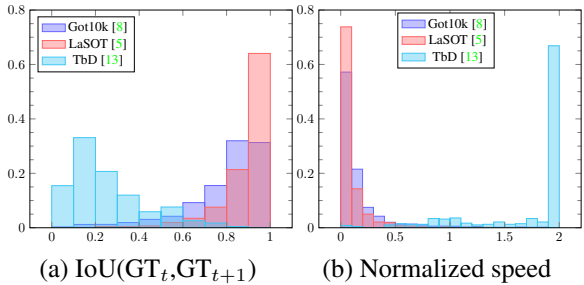

FMOs were introduced in~\cite{fmo}, where the authors created the first dataset containing fast moving objects and proposed a proof-of-concept algorithm to detect and track such objects.
The problem of detecting and tracking FMOs has been unnoticed by the research community, and such objects are not present in standard tracking datasets such as VOT~\cite{vot16}, OTB~\cite{otb}, and ALOV~\cite{alov}. 
More precisely, they may or may not be present, but FMO is never the object intended for tracking by the ground truth annotation.
We highlight this by running FMO retrieval by the proposed method and detect FMOs in very recent standard large datasets like GOT-10k~\cite{got10k} and LaSOT~\cite{lasot} and find unannotated FMOs (Fig.~\ref{fig:intro}, bottom). 
By processing the ground truth bounding boxes in more than 5M frames from both datasets, we noticed that there is a significant overlap between consecutive bounding boxes in almost all cases, and the speed is close to zero (Fig.~\ref{fig:hist}). 
In contrast, the recently proposed TbD dataset~\cite{tbd} with fast moving objects has almost zero intersection over union between consecutive objects, and the speed is higher.

Apart from~\cite{fmo}, there are a few methods that deal with FMOs.
Tracking by Deblatting (TbD)~\cite{tbd} jointly solves an inverse problem of \textit{deblatting} (\textit{debl}urring and m\textit{atting}) and decouples the input images into trajectory, appearance, and mask. 
All previous methods are based on classical energy minimization. 
While powerful, they still require many parameter tuning and are slow. 
In most cases, they have to be tuned for different sequences separately. 
Also, their performance is not entirely satisfactory, and they force several limiting assumptions,~\eg static camera, high contrast between the object and the background, and spherical shape. 

This paper proposes a novel method for FMO detection and trajectory and appearance estimation using a combination of deep learning and energy minimization based deblurring. 
Instead of solving an ill-posed and complex deblatting~\cite{tbd} problem, we first solve the matting problem that separates the background from the input image by learning from synthetic data. 
Then, an easier deblurring problem is solved by energy minimization.
The proposed FMODetect is real-time capable and achieves state-of-the-art performance, which allows us to tackle challenging applications such as FMO retrieval and tracking in long videos (Fig.~\ref{fig:retrieval}). 

In summary, we make the following \textit{contributions}:
\begin{itemize}
    \item We present the first learned approach to the problem of FMO detection. Compared to the previous methods, the proposed one is simpler, does not require extensive parameter tuning, and works with the same settings for a wide range of scenarios,~\eg objects of different complexities, moving among various trajectories, and captured by moving or zooming cameras. The method is trained on a new synthetic dataset and generalizes well to unseen and more difficult real-world data.
    \item By separating deblatting into matting and deblurring, computational and problem complexities are simplified, and the method achieves state-of-the-art results. Our method is an order of magnitude faster than the previous methods, and the detection part is real-time capable. This makes applications such as robust FMO retrieval from large video collections realistic.
    \item For the sharp appearance estimation, we propose a novel energy minimization-based deblurring. 
\end{itemize}

\section{Related work}
The pioneering work in~\cite{fmo} introduced FMOs but was limited by many assumptions such as a linear trajectory, high contrast, and no occlusions. Some of them have been addressed in the follow-up method called Tracking by Deblatting (TbD)~\cite{tbd}. TbD solves an inverse problem of \textit{debl}urring and m\textit{atting} by alternating iterative minimization, followed by fitting piecewise parabolic curves. As a by-product, TbD estimates sharp 2D object appearance, which is used for long-term tracking of fast moving objects. Both methods introduced datasets with fast moving objects. The ground truth in the FMO dataset~\cite{fmo} was only a binary mask with rough locations of FMOs in each frame. The TbD dataset~\cite{tbd} provides more ground truth information from a high-speed camera, such as object templates and sub-frame object trajectories.  

The above-mentioned methods assumed causal processing of video frames, which means that the trajectory at the current frame is estimated using only information from previous frames. However, some scenarios involving fast moving objects do not require causal processing, and some latency is acceptable. Non-causal Tracking by Deblatting (TbD-NC)~\cite{tbdnc} is the first method to exploit this fact. 
TbD-NC takes detections in each frame from a causal method as input and produces a single trajectory for the whole sequence. 
Then, the final trajectory is a continuous, piecewise polynomial curve. TbD-NC improves the accuracy and recall of trajectory estimation of fast moving objects. Also, it allows for accurate speed estimation, which is on par with expensive radar guns used in professional tennis matches. Applications such as temporal super-resolution become significantly better with non-causal estimation.

The method proposed in this paper is causal, and computations are done for each frame independently. Therefore, the method in TbD-NC can be applied to improve the trajectories estimated by the proposed method.

The authors in~\cite{fmo} introduced fast moving objects together with the corresponding image formation model 
\begin{equation}
	\label{eq:model}
	I = H*F + (1-H*M)B,
\end{equation}
where the image is formed as a mixture of blurred object appearance and the background. Motion blur is modeled by the convolution of the object trajectory, represented by the blur kernel $H$, and the sharp object appearance $F$. 
The influence of background depends on the indicator function $M$ of appearance $F$. For more details, the reader is referred to~\cite{fmo}.

Many deblurring methods have been proposed. They all assume small motions like in~\cite{Jin_2018_CVPR,nah,Purohit_2019_CVPR}. The method in~\cite{Pan_2019_CVPR} considers larger motions but only in a trivial case.
We tried several of these methods, and none was able to deblur or reconstruct fast moving objects. 
Special deblurring of fast moving objects is also proposed in~\cite{kotera2020, kotera, sroubek2020}, but only a simple case is studied. 
The only competitive method is TbD~\cite{tbd}. 
TbD-3D~\cite{tbd3d} is an extension towards 3D objects with 3D rotations, which can be directly applied to the output of the proposed method. 

Network architectures similar to the proposed one have been used in other methods. 
Hou~\etal~\cite{Hou_2019_ICCV} proposed a two-encoder-two-decoder architecture but matting of sharp images is considered, as well as in~\cite{deep_matting,late_fusion}. 
The architecture in~\cite{motion_blur} is similar, but the task is inverse~--~to synthesize the motion blur given sharp images. 
Similar U-Net~\cite{unet} architectures have been applied in many image-to-image translation tasks~\cite{10.5555/3326943.3327062,8100115}.

\begin{figure*}[t]
\centering
\includegraphics[width=0.99\textwidth]{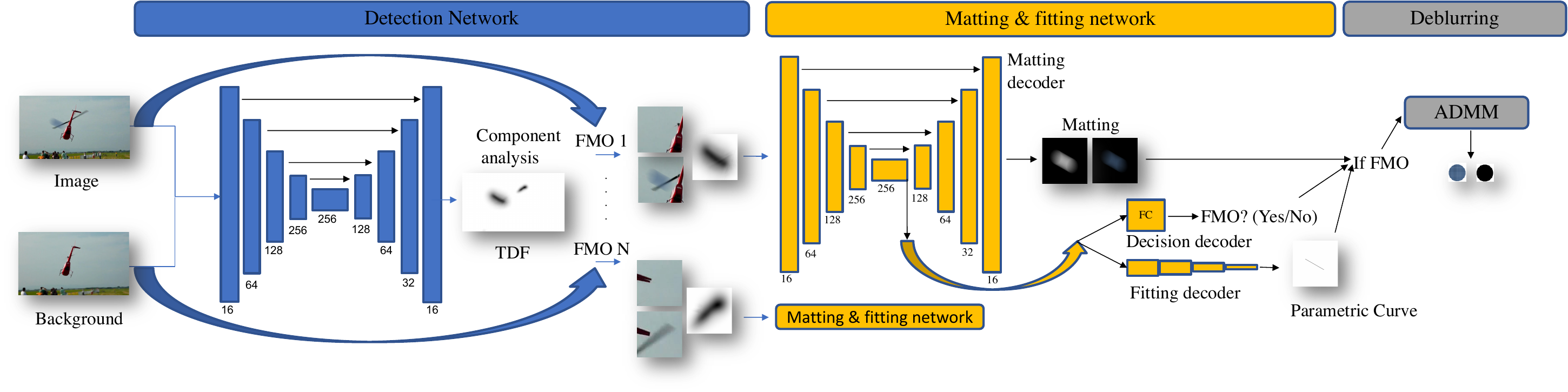}
\vspace*{-1em}
\caption{The FMO detection network recovers all fast moving objects (FMOs) in the input image. Then, the matting and fitting network is applied on each detected FMO. The outputs, blurred object without background and its trajectory, are then further used in the proposed deblurring~\eqref{eq:deconv} to establish sharp object appearance and refine the trajectory.}
\label{fig:arch}
\end{figure*}

\section{Method}
%
The proposed method detects all fast moving objects in an image, recovers their trajectory, and removes the influence of the background,~\ie matting. Then an optional step is applied to get sharp object appearance, based on energy minimization based deblurring. An overview of the architecture is shown in Fig.~\ref{fig:arch}.

First, an FMO detection network proposes potential regions with fast moving objects using a truncated distance function (TDF) to represent the trajectory.
Second, for each region found by the detection network, the matting and fitting network estimates the parametric curve describing FMO trajectory and performs matting~--~removing the background from the input image and leaving only the blurred object appearance $H*F$ and $H*M$ from image formation model~\eqref{eq:model}.
This representation can be seen as an FMO moving in front of a completely black background. 
Estimating just $H*F$ is not sufficient since the shape of the object should be estimated besides $F$, which models only the surface color. 
Consider an object of black color; here $H*M$, unlike $H*F$, is non-zero. 
The final step,  deblurring, is done by energy minimization similar to deblatting~\cite{tbd}. 
However, by first removing the blended background, it is easier and faster. 

\subsection{Detection network}
The inputs to the detection network are the input image $I$ and the background $B$, estimated as a median of the last three frames in a video sequence.
The median image approximation worked best in our experiments. 
Changing the number of frames for the background calculation is essentially a trade-off between sensitivity to camera motion (lower number is better) and noise reduction (higher is better). 
Our experiments showed that using 3 frames was the most robust setting for camera-motion which provided a sufficiently clean background.

The output is the truncated distance function $D$, measuring the distance to the object trajectory. If we assume that pixels are in some domain $\mathbb{X} \subset \mathbb{R}^2$, then for each pixel $x \in \mathbb{X}$, $D$ is defined as

\begin{equation}
	\label{eq:dt}
	D(x) = 1 - \min\left(1, \min_t \frac{||x - C(t)|| _2}{2r}\right),
\end{equation}
where $C(t) : [0,1] \to \mathbb{R}^2 $ is the ground truth continuous trajectory, and $r$ is the ground truth object radius.

The ground truth for the training sequences is prepared as follows.  
First, a distance function to the trajectory is computed. 
Then, it is divided by the object diameter~--~the maximum distance between any two points on the object. 
Note that pixels influenced by the object are located at half diameter distance to the trajectory. 
However, the network converges much faster when a larger neighborhood is considered.  
Experiments confirmed that it is beneficial to invert the TDF output,~\ie 0 means maximum distance (the truncation level), and 1 means 'on the trajectory.' 
The network benefits from the fact that the output is expected to be zero at most locations, which is easier to predict.


The architecture is similar to U-Net~\cite{unet}, and it is fully convolutional. 
The encoder contains four blocks as in~\cite{motion_blur}, each followed by a max-pooling layer. 
Each block is built from three convolution layers with filter size $3 \times 3$, followed by a Leaky ReLU~\cite{Maas13rectifiernonlinearities} with slope 0.1. 
The decoder part also contains four blocks, each followed by a transposed convolution layer~\cite{deconv,deconv2}. 
Inputs and outputs are rescaled to the resolution of $256 \times 512$ pixels, with zero mean and unit variance. The full-resolution images are used for testing. 

The loss function for the detection network is defined as
\begin{equation}
	\label{eq:loss_det}
	\begin{split}
	\mathcal{L}_d = \frac{1}{N_1}\sum_{D(x) > 0} \|D(x) - \hat{D}(x)\|_1 + \frac{1}{N_0} \sum_{D(x) = 0} \|\hat{D}(x)\|_1, \\
	\end{split}
\end{equation}
where $D$ corresponds to the ground truth TDF, and $\hat{D}$ to the estimated. $N_1$ is the number of positive pixels in the ground truth TDF, and $N_0$ is the number of zero value pixels where a fast moving object is not present in the neighborhood. 

\newcommand{\AddName}{\raisebox{2.7em}{\rotatebox[origin=c]{270}{\scriptsize GT \quad \quad \quad Ours}}}
\begin{figure*}[t]
\centering
\footnotesize
\setlength{\tabcolsep}{0.05em}
\begin{tabular}{@{}cc@{}c@{}c@{}c@{}c@{}}

\includegraphics[height=0.135\textwidth]{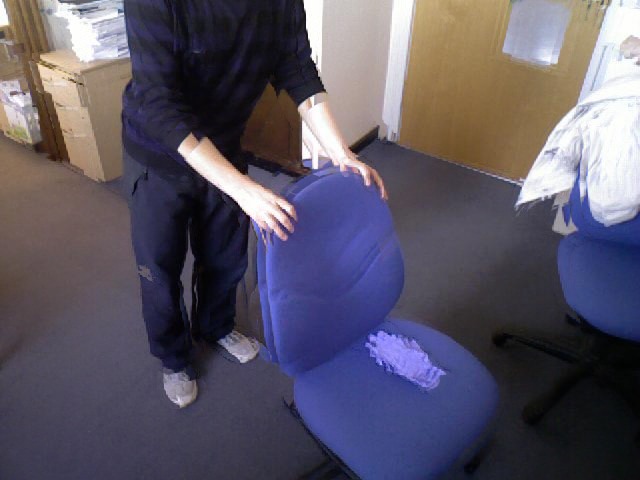} & 
\includegraphics[height=0.135\textwidth]{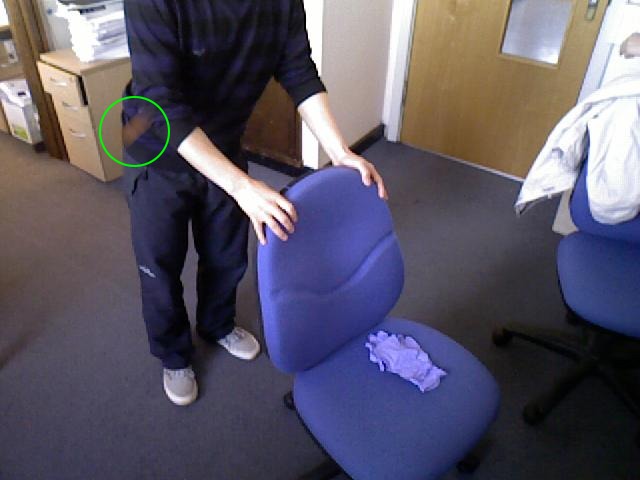} &
\includegraphics[height=0.135\textwidth,frame]{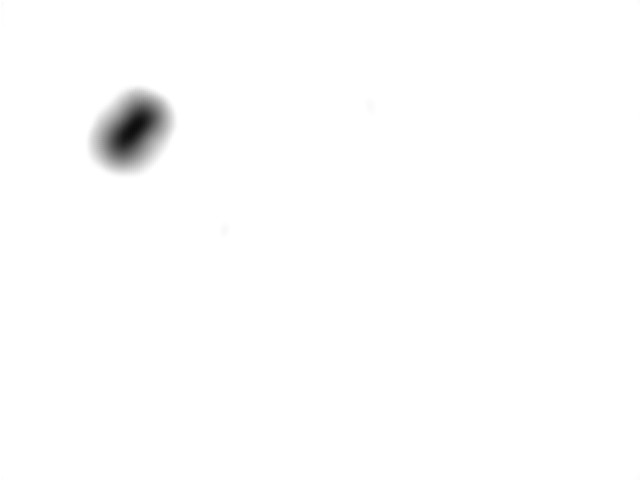} &
\includegraphics[height=0.135\textwidth]{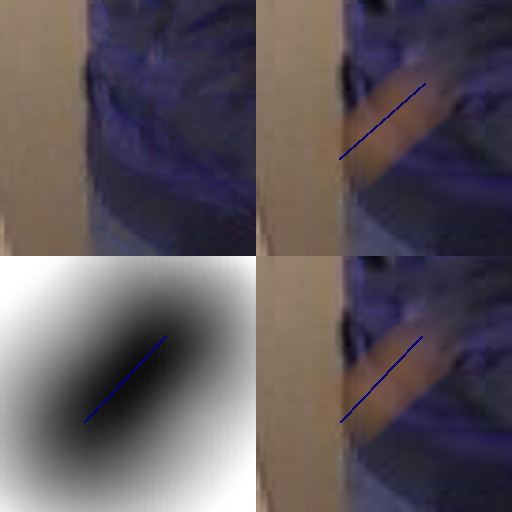} &
\includegraphics[height=0.135\textwidth]{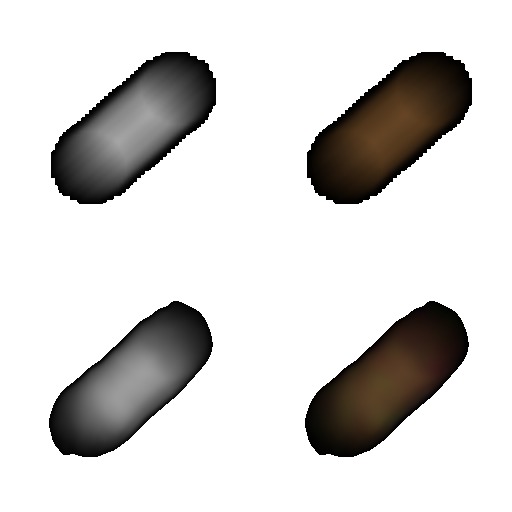} & \AddName \\

\includegraphics[height=0.135\textwidth]{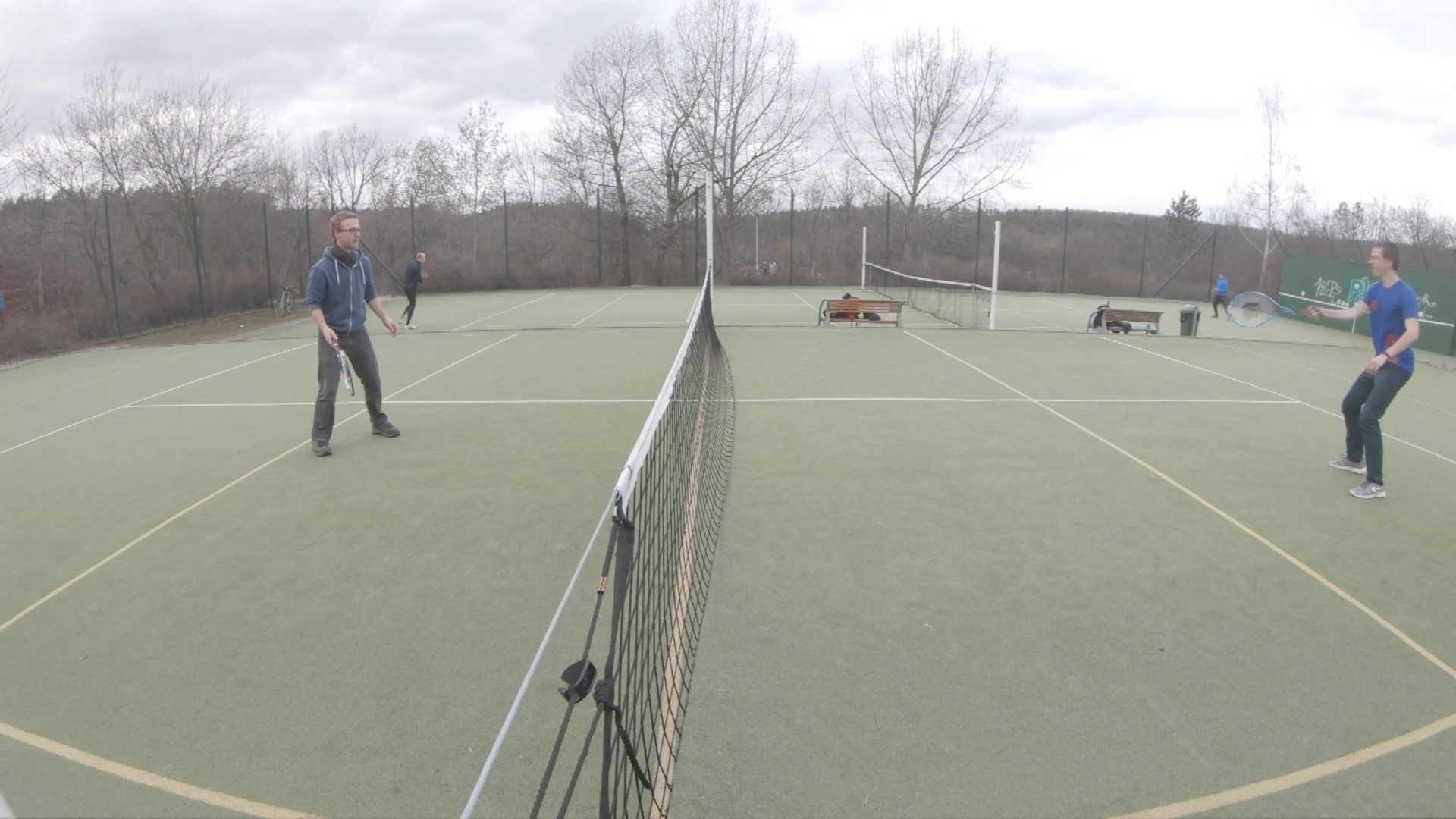} & 
\includegraphics[height=0.135\textwidth]{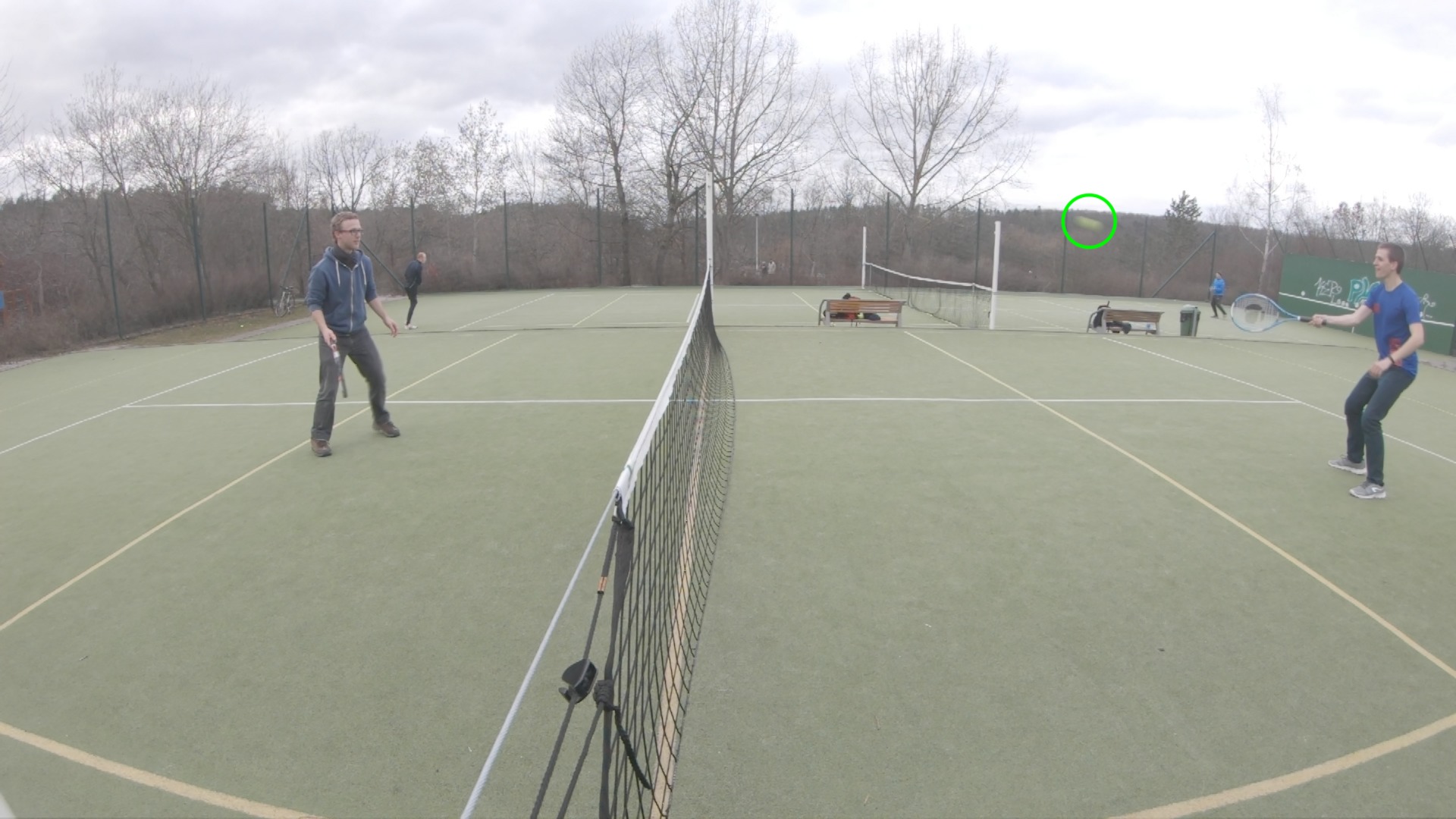} &
\includegraphics[height=0.135\textwidth,frame]{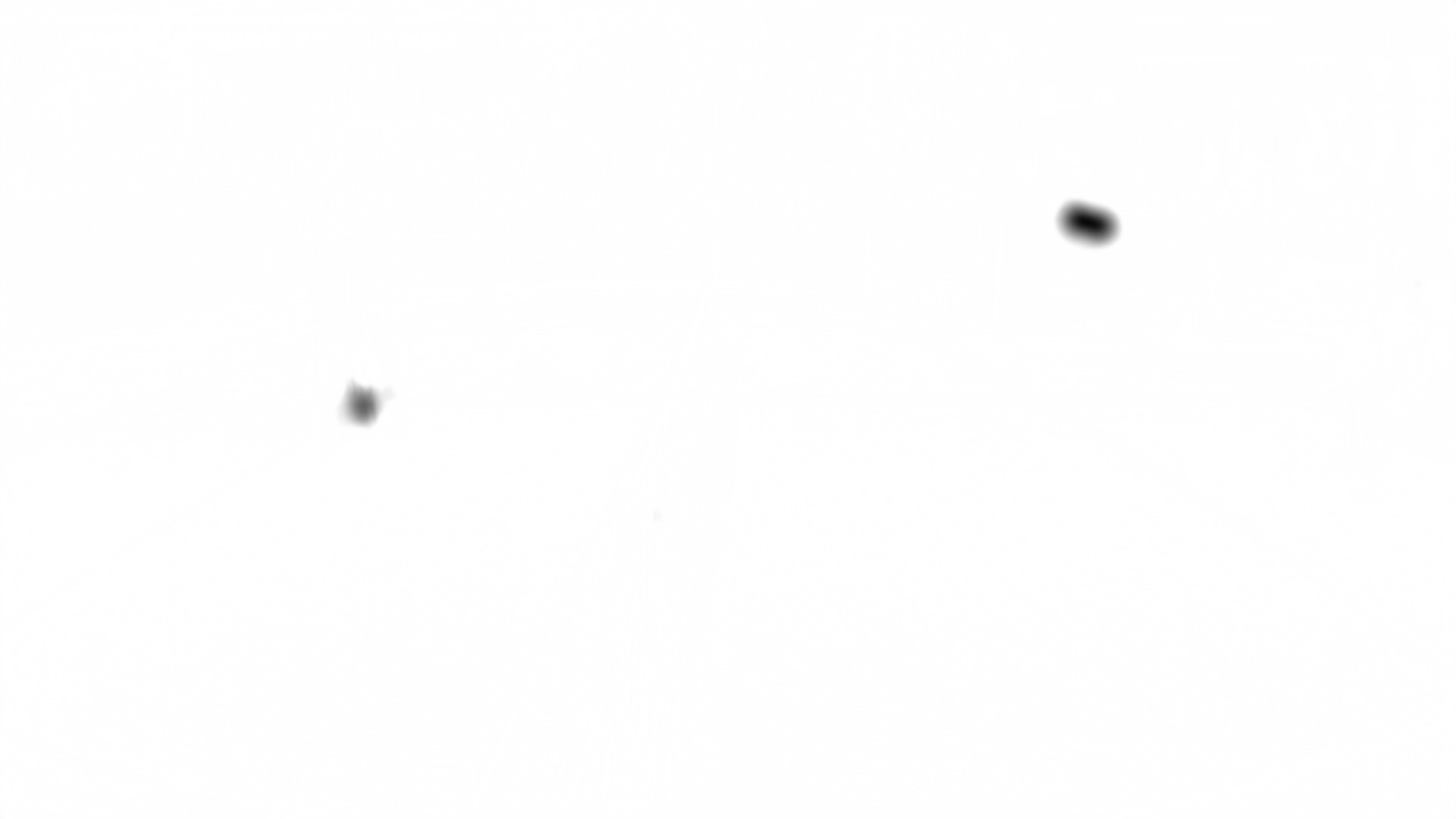} &
\includegraphics[height=0.135\textwidth]{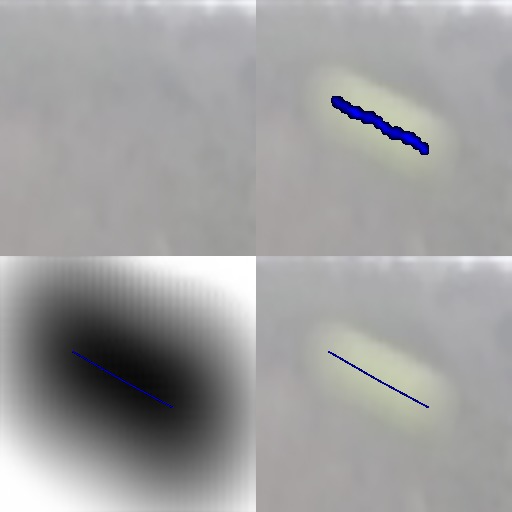} &
\includegraphics[height=0.135\textwidth]{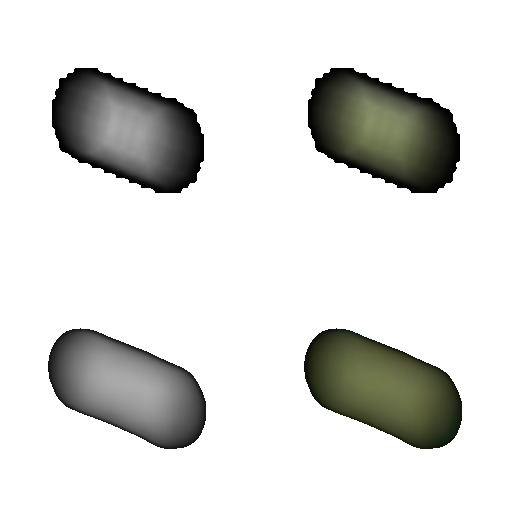}  & \AddName  \\

Background & Image & TDF & Trajectory & Matting \\

\end{tabular}

\caption{Step by step data processing by our method. We include frames from the synthetic dataset (top row) and real-world TbD dataset~\cite{tbd} (bottom row). \textit{Truncated distance function} (TDF) is an intermediate output of our method for a detected trajectory. Trajectory reconstruction is plotted on the crops. The detected object is marked by a green circle. Examples of deblurring after matting are in Fig.~\ref{fig:res_deblur}. }
\label{fig:res_pipeline}
\end{figure*}

\subsection{Matting and fitting network}
The output $D$ of the detection network is binarized by the threshold value $\epsilon$, and all connected components are extracted. For each connected component, we take crops from the input image, the background, and the non-binarized TDF. 
They are all scaled to the same resolution of $256 \times 256$ pixels and are inputs to the network. 
The matting and fitting network output is three-fold: blurred object appearance without the influence of the background (matting), parametric trajectory, and a binary decision whether the crop corresponds to an FMO. 
The last binary output is used to correct the mistakes of the detection network. 
Experimentally, we have confirmed that the binary predictor removes $\approx\!\!70\%$ of false positives (measured on the TbD dataset~\cite{tbd}).

The matting and fitting network has a one-encoder-three-decoder architecture. 
The architecture is similar to the detection network. 
The transposed convolution operators are replaced by upsampling operators to avoid checkerboard artifact~\cite{checkerboard}. 
The output of the matting decoder is a 4-channel image ($H*F$, $H*M$) since $F$ is an RGB image and $M$ is a gray-scale image. The fitting decoder is a multi-layer fully connected network with the final output dimension of 8 channels: the degree of freedom of the estimated parametric trajectory. 
The last decoder contains only one fully connected layer attached to the latent space and is followed by a sigmoid activation with a single output.

We assume that the trajectory within a frame is one of the following: a line, a parabola, a piece-wise line with one bounce. 
Then, a common parametric representation of these three cases is
\begin{equation}
	\label{eq:traj}
	\begin{split}
	\mathcal{C}(t) = c_0 + c_1 \text{min}(2t,1) + c_2 \text{min}(2t,1)^2 + \\
	   + c_3 \text{max}(2t-1,0), \qquad \text{s.t.} \quad 0\leq t \leq 1,
	\end{split}
\end{equation}
where parameters $c_k \in \mathbb{R}^2$. 
In case of a line, parameters $c_2$ and $c_3$ are equal to zero vectors. 
In case of a parabola, only $c_3$ is zero, and for a piece-wise line $c_2$ is equal to zero. 
This parametrization can also represent a parabolic curve with a linear tail. However, to avoid this case, whenever $\| c_3 \| > 1$, we set $c_2$ to zero, otherwise $c_3$ is set to zero.

The loss function of the matting and fitting network is
\begin{equation}
	\label{eq:loss}
	\begin{split}
	\mathcal{L} = \alpha_{a} b(\|H*F - \hat{H}_{F}\|_1 + \|H*M  - \hat{H}_{M}\|_1) + \\
	 + \alpha_{c} b \mathcal{L}_c (\mathcal{C},\hat{\mathcal{C}})  + \alpha_{b} \text{BCE}(b,\hat{b}), \\
	\end{split}
\end{equation}
which combines three terms. 
The first one corresponds to blurred appearance reconstruction represented by the ground-truth pair ($H*F$, $H*M$). 
Estimated blurred appearance without background is denoted by ($\hat{H}_F$, $\hat{H}_M$). 
The second term penalizes deviations from the ground truth trajectory by the curve loss $\mathcal{L}_c$ that samples the trajectory at three points and computes an average distance between each pair. 
Since the direction is not known, we compute both options and choose the one with the minimal distance. 
The last term computes binary cross entropy (BCE) between the estimated and the ground truth binary variable $b$ that indicates whether the crop contains an FMO. 
All previous terms are multiplied by the ground truth binary variable to allow the network to output the best guess of trajectory and appearance even when there is no fast moving object, penalizing only for the incorrectly estimated binary variable $b$. 

\subsection{Deblurring}
We formulate deblurring as an optimization problem. The output $\hat{\mathcal{C}}$ of the fitting decoder is used for initialization.
We also include optimization with respect to $H$ in deblurring to account for small inaccuracies in the blur kernel. 
The object appearance and mask are then estimated by solving
\begin{equation}
	\label{eq:deconv}
	\begin{split}
	\min_{F,M,H} \frac{1}{2}\left( \|H*F-\hat{H}_{F}\|_2^2 + \|H*M-\hat{H}_{M}\|_2^2\right) + \\
	+ \alpha_F\|\nabla F\|_1 + \alpha_M\|\nabla M\|_1 \\
	\end{split}
\end{equation}
s.t. $0\leq F \leq M \leq 1$, $H \geq 0$ and $\sum_i H_i = 1$. The deblurring inputs are $\hat{H}_F$ and $\hat{H}_M$, which are the outputs of the matting decoder. All constraints are convex sets and we solve the problem using the alternating direction method of multipliers \cite{Boyd2011}. The variables are split into blur kernel $H$ and appearance model $(F,M)$, and both are updated iteratively. In case a parametric curve from $H$ is needed,~\eg for evaluation, we run curve fitting from~\cite{tbd}.

The deblurring problem~\eqref{eq:deconv} is much easier to solve than the deblatting problem~\cite{tbd}. Deblurring provides more accurate results, as illustrated in several examples in Fig.~\ref{fig:res_deblur}. It is twice as fast as deblatting, and if minimization with respect to $H$ is omitted from \eqref{eq:deconv}, it is 10 times faster. 

Among other limitations of deblatting, which are solved by the proposed approach, are also failures to separate the background or a shadow correctly (Fig.~\ref{fig:res_deblur}, golf) or to deal with low contrast objects (Fig.~\ref{fig:res_deblur}, softball).


\newcommand{\MySpace}{\hspace{0.7em}}
\newcommand{\AddNameD}[1]{\raisebox{0em}{\rotatebox[origin=c]{270}{#1}}}

\newcommand{\addimg}[1]{\includegraphics[height=0.156\textwidth]{imgs/deblur/#1}}

\newcommand{\AddOne}[1]{\begin{tabular}{@{}cccc@{}}
\addimg{#1/inp.jpg} & 
\addimg{#1/traj.jpg} & 
\addimg{#1/fm.jpg} &
\addimg{#1/deblur.jpg} \\
\end{tabular}}

\begin{figure*}[t]
\footnotesize
\centering
\setlength{\tabcolsep}{0em}

\vspace*{-2em}
\begin{tabular}{@{}cccccc@{\hskip 0.1em}c@{}}
\rotatebox[origin=c]{270}{\AddOne{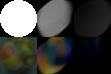}} &
\rotatebox[origin=c]{270}{\AddOne{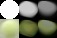}} &
\rotatebox[origin=c]{270}{\AddOne{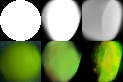}} &
\rotatebox[origin=c]{270}{\AddOne{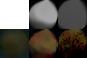}} &
\rotatebox[origin=c]{270}{\AddOne{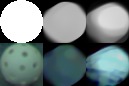}} &
\rotatebox[origin=c]{270}{\AddOne{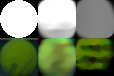}} & \AddNameD{\scriptsize \quad  \qquad (a) Inputs \qquad (b) Traj. \qquad (c) Matt. \qquad (d) GT \qquad (e) Ours \qquad (f) TbD~\cite{tbd} \qquad \quad} \\[-2.5em]
Softball & Tennis & Golf & Cube & Floorball & Tennis \\
\end{tabular}
\caption{Matting and deblurring of fast moving objects in the TbD dataset~\cite{tbd} using the proposed method, compared to deblatting~\cite{tbd}. From top to bottom: (a) input image and background by median, (b) left: estimated trajectory plotted on the input image, right: estimated truncated distance function (TDF), (c) output of the matting decoder, (d) ground truth (GT), (e) the proposed deblurring, (f) deblatting~\cite{tbd}. }
\label{fig:res_deblur}
\end{figure*}

\subsection{Training}
The standard datasets with FMOs are the TbD~\cite{tbd} and FMO~\cite{fmo} datasets, with 12 and 16 sequences respectively. 
Ground truth trajectories are estimated by running a state-of-the-art tracking algorithm on high-speed camera footage where the blur is not present. 
Since the ground truth of matting is not available and is not realistic to get, we created a new synthetic dataset with fast moving objects, called VOT-FMO. 
The VOT-FMO dataset is based on the VOT dataset~\cite{vot18}~--~a standard dataset for tracking. 
As previously shown in~\cite{fmo} and Fig.~\ref{fig:intro}, standard tracking datasets do not label FMOs, even though they still may be present there. 
Thus, the synthetic dataset is created by applying the FMO formation model~\eqref{eq:model} with backgrounds taken from the VOT dataset and with artificially inserted FMOs. 
Artificial FMOs are random discs of up to 100 pixels in radius with textures generated by GeoPatterns\footnote{\href{https://github.com/jasonlong/geo_pattern}{https://github.com/jasonlong/geo\_pattern}}. 
Training with objects of various sizes results in good scale invariance.
We generate 5000 random frames for training and 500 for validation. 

Weights for the loss are set empirically: $\alpha_{a} = 15$, $\alpha_{b} = 0.4$. 
Parameter $\alpha_{c}$ is set to $\frac{4}{256}$, which normalizes the curve loss to produce one unit when the average distance between trajectories is $\frac{1}{4}$ of the crop width of 256 pixels. 
Regularization weights for deblurring were experimentally set to $\alpha_F = 0.001$ and $\alpha_M = 0.05$.
We found that the value of the binarization threshold $\epsilon$ between 0.1 and 0.4 gives similar results. For experiments, we set it to 0.3.

The network is trained by minimizing the joint loss function using Adam optimizer~\cite{adam} with learning rate $2 \times 10^{-5}$. 
The implementation is written in Python using Keras~\cite{chollet2015} framework with Tensorflow~\cite{tensorflow2015-whitepaper} backend. 
The detection network contains 4.8M parameters, and the matting and fitting network 5.7M parameters. 
Training takes around three days on Nvidia GTX 1080 Ti GPU with 11 GB memory. 
Inference takes on average 0.05 seconds on a single image of 600 by 960 pixels, which shows the real-time capability of the proposed method. 
In comparison, the previous methods~\cite{tbd,tbdnc,fmo} require a couple of seconds per frame.

\begin{figure}[t]
\centering
\scriptsize
\newcommand{\sz}{0.107}
\newcommand{\szz}{0.08}
\newcommand{\szzz}{0.06}
\setlength{\tabcolsep}{0.1pt}
\begin{tabular}{ccccccccc}

\includegraphics[height=\sz\textwidth]{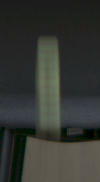} & 
\includegraphics[height=\sz\textwidth]{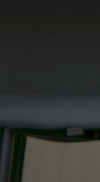} & 
\includegraphics[height=\sz\textwidth]{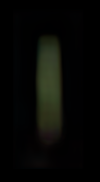} & 
\includegraphics[height=\sz\textwidth]{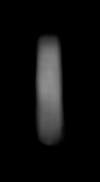} & 
\includegraphics[height=\sz\textwidth]{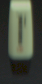} & 
\includegraphics[height=\sz\textwidth]{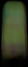} & 
\includegraphics[height=\sz\textwidth]{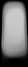} & 
\includegraphics[height=\sz\textwidth]{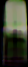} & 
\includegraphics[height=\sz\textwidth]{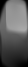} \\

$I$ & $B$ &  $H_F$ & $H_M$ & GT & $F$ & $M$ & $F$~\cite{tbd} & $M$~\cite{tbd} \\

\end{tabular}

\caption{Additional deblurring results on a non-spherical object. From left to right: input image $I$, background $B$, estimated $H_F$ and $H_M$ by the matting decoder (removed background), the ground truth from a high-speed camera, object appearance $F$ and mask $M$ estimated by the proposed deblurring, and compared to reconstructions by the deblatting method~\cite{tbd}. 
}
\label{fig:falling}
\end{figure}

Fig.~\ref{fig:res_pipeline} shows step-by-step visualizations of method outputs. Deblurring results are shown in Fig.~\ref{fig:res_deblur} and Fig.~\ref{fig:falling}.

\definecolor{maxClr}{rgb}{0,0.5,1}
\definecolor{max2Clr}{rgb}{1,0.1,0.6}
\definecolor{darkblue}{rgb}{0.0, 0.0, 0.55}
\definecolor{darkcerulean}{rgb}{0.03, 0.27, 0.49}
\definecolor{darkpastelblue}{rgb}{0.47, 0.62, 0.8}

\begin{table*}[t]
\centering
\setlength{\tabcolsep}{3mm}
\newcommand{\winner}[1]{\textcolor{black}{\textbf{#1}}}
\begin{center}
\begin{tabular}{lcccccc}

\specialrule{.1em}{.05em}{.05em}


  & \multicolumn{4}{c}{Causal processing} & \multicolumn{2}{c}{Non-causal processing} \\
 	\cmidrule(lr){2-5} \cmidrule(lr){6-7}

 &  FMO~\cite{fmo}  & TbD~\cite{tbd} & (a) FMODetect  & (b) + deblurring & (c) + NC  & TbD-NC~\cite{tbdnc} \\ 

 \midrule 

Recall$\uparrow$ & 0.56 & 0.96  & \winner{0.97} & \winner{0.97} & {0.99} & {0.99} \\

TIoU$\uparrow$ & 0.352 & 0.713  & 0.519 & \winner{0.715}  &  {0.781} & 0.779 \\

Runtime [1/s]$\uparrow$ & 1 fps & 0.2 fps& \winner{20 fps} & 0.4 fps & N/A  & N/A \\

\specialrule{.1em}{.05em}{.05em}

\end{tabular}
\end{center}
\vspace*{-0.8em}
\caption{Recall and trajectory accuracy TIoU~\eqref{eq:tiou} on the TbD dataset~--~comparison of the proposed methods to: FMO method~\cite{fmo}, TbD~\cite{tbd}, and TbD-NC~\cite{tbdnc} methods. Versions of the proposed method: (a) real-time with trajectories estimated by the network, (b) with the proposed deblurring, and (c) with non-causal post-processing from~\cite{tbdnc}. The best performing causal method is highlighted in \textbf{bold}. Non-causal methods require expensive processing of all frames simultaneously. Qualitative comparison is in Fig.~\ref{fig:res_tbd}.}
\label{tbl:tbd}
\end{table*}

\begin{table}[t]
\centering
\begin{center}
\setlength{\tabcolsep}{1mm}
\begin{tabular}{llccc}
\toprule
Dataset & Metric  &  FMO & TbD  & FMODetect \\  

\midrule

 \multirow{2}{*}{FMO~\cite{fmo}} & Recall$\uparrow$ &  35.5  & 41.1 &  \textbf{48.8} \\
 & Precision$\uparrow$ &  59.2 & 81.6 &  \textbf{84.9} \\
\cmidrule(lr){2-5}
 \multirow{2}{*}{TbD-3D~\cite{tbd}} & Recall$\uparrow$ & 63.8 & 69.8 & \textbf{83.5} \\
 & Precision$\uparrow$ & 81.7 & \textbf{98.2} & 87.9 \\
\cmidrule(lr){2-5}
 \multirow{2}{*}{Falling~\cite{kotera2020}} & Recall$\uparrow$  & 51.5 & 77.2 & \textbf{82.0} \\
 & Precision$\uparrow$ & 84.0 & 92.2 & \textbf{98.9} \\
\cmidrule(lr){2-5}
 YouTube & Recall$\uparrow$ & 36.4 & 27.3 & \textbf{90.1} \\  
\midrule
  & Runtime [1/s]$\uparrow$ & 1 fps & 0.2 fps & \textbf{20 fps} \\  
\bottomrule
\end{tabular}
\end{center}
\vspace*{-0.8em}
\caption{Precision and recall, averaged on the FMO~\cite{fmo}, TbD-3D~\cite{tbd}, and the falling objects~\cite{kotera2020} datasets. We additionally report recall of detected fast moving objects in a trick shots video (Fig.~\ref{fig:retrieval}) with manually annotated ground truth. FMODetect is compared to the TbD~\cite{tbd} and FMO~\cite{fmo} methods. The best performing method is highlighted in bold.
}
\label{tbl:datasets}
\end{table}

\section{Experiments}
We evaluate the FMO detection accuracy on several real-world datasets. 
Detection accuracy is measured by precision and recall. 
A detection is called true positive if intersection over union between the predicted region influenced by the FMO and the ground truth one is greater than 0.1. 
Then, precision is the percentage of true positives from all detections.
Recall is defined as the ratio between the number of true positives and the number of ground truth annotations.
If the ground truth sub-frame trajectory $\mathcal{C}^*$ is available, we measure the accuracy of the estimated trajectory $\mathcal{C}$ by Trajectory Intersection over Union (TIoU)~\cite{tbd}:
\begin{equation}
	\label{eq:tiou}
	\operatorname{TIoU}(\mathcal{C},\mathcal{C}^*) = \int_{t} \operatorname{IoU}\left(\rule{0pt}{2ex}M^*_{\mathcal{C}(t)},\, M^*_{\mathcal{C}^*(t)}\right)\mathrm{d}t,
\end{equation}
where object mask is placed on a trajectory point $\mathcal{C}(t)$, denoted by $M^*_{\mathcal{C}(t)}$, and the standard Intersection over Union (IoU) is computed. Then the average is calculated along the whole trajectory.
We compare to the FMO~\cite{fmo} and TbD~\cite{tbd} methods, both designed to detect and track FMOs.

\subsection{Ablation study}
Ablation study is performed on the real-world TbD dataset~\cite{tbd}. 
The first ablated version is fully learning-based prediction.
As shown in Table~\ref{tbl:tbd} (a), FMODetect is very accurate in terms of FMO detection (measured by recall). 
However, if we further evaluate trajectories predicted by the network, the trajectory accuracy (TIoU) is not on par with the state-of-the-art.
If we instead of solving the deblatting problem jointly by the network (Table~\ref{tbl:tbd} (a), TIoU) solve only the matting problem by the network and run the proposed deblurring to further refine the trajectories (Table~\ref{tbl:tbd} (b)), we achieve state-of-the-art trajectory accuracy.

We also show that non-causal post-processing~\cite{tbdnc} can complete missing detections and make trajectories more consistent within the whole sequence. Compared to TbD-NC~\cite{tbdnc}, TIoU is marginally better.

\subsection{Detection accuracy}
The TbD dataset captures simplistic scenario of spherical almost uniform objects moving in a constant scene. 
In this case, the difference between FMODetect and the other methods is marginal.
Further experiments demonstrate that in more challenging cases, the proposed method is significantly more robust. 
We evaluate detection accuracy on the FMO dataset~\cite{fmo} (various small objects), TbD-3D dataset~\cite{tbd3d} (spherical highly textured objects moving in 3D), and the falling objects dataset~\cite{kotera2020} (arbitrarily shaped objects,~\eg key, pen, marker). 
Additionally, we gather 11 short sub-videos from YouTube trick shots video (some are in Fig.~\ref{fig:retrieval}) and manually annotate the object of interest. 
Since other FMOs are not annotated in the YouTube dataset, we measure only recall here.
In terms of detection accuracy, FMODetect outperforms all other methods by wide margin on all datasets (Table~\ref{tbl:datasets}).
Only on the TbD-3D dataset, the precision is lower since our method also detected the shadow of the object, and it is still questionable if such detection should be considered a correct detection.
Compared to other methods, FMODetect is faster and runs in real-time.

\begin{figure*}[t]
\centering
\newcommand{\sz}{0.148}
\setlength{\tabcolsep}{0pt}
\begin{tabular}{@{}c@{}c@{}c@{}c@{}c@{}}

\rotatebox{90}{\hspace{20pt}FMODetect} &
\resizebox {!}{\sz\textwidth} {\resizebox {!}{0.2\textwidth} {\begin{tikzpicture} 
\begin{axis}[y dir=reverse, 
 xmin=1,xmax=960, 
 ymin=1,ymax=600, 
 xticklabels = \empty, yticklabels = \empty, 
 grid=none, axis equal image] 
\addplot graphics[xmin=1,xmax=960,ymin=1,ymax=600] {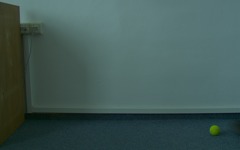}; 
\addplot [,domain=0:1,samples=2,style=semithick,color=green]({705.5965 + 66.5672*x},{538.2216 + -40.5032*x});  
\addplot [,domain=0:1,samples=2,style=semithick,color=green]({648.4883 + 26.2668*x},{517.6863 + 4.8729*x});  
\addplot [,domain=0:1,samples=2,style=semithick,color=green]({674.7551 + 31.0134*x},{522.5592 + 4.3764*x});  
\addplot [,domain=0:1,samples=2,style=semithick,color=green]({588.1389 + 24.5226*x},{516.9293 + 4.8709*x});  
\addplot [,domain=0:1,samples=2,style=semithick,color=green]({612.6615 + 30.1963*x},{521.8002 + 4.4004*x});  
\addplot [,domain=0:1,samples=2,style=semithick,color=green]({526.0557 + 25.5664*x},{517.4326 + 3.6837*x});  
\addplot [,domain=0:1,samples=2,style=semithick,color=green]({551.6221 + 30.9477*x},{521.1162 + 3.2324*x});  
\addplot [,domain=0:1,samples=2,style=semithick,color=green]({467.5032 + 24.0335*x},{516.7332 + 4.186*x});  
\addplot [,domain=0:1,samples=2,style=semithick,color=green]({491.5367 + 29.4524*x},{520.9192 + 3.8128*x});  
\addplot [,domain=0:1,samples=2,style=semithick,color=green]({406.9511 + 24.7502*x},{514.1136 + 4.6589*x});  
\addplot [,domain=0:1,samples=2,style=semithick,color=green]({431.7014 + 29.8666*x},{518.7725 + 4.0825*x});  
\addplot [,domain=0:1,samples=2,style=semithick,color=green]({346.0869 + 26.7206*x},{513.4036 + 4.1245*x});  
\addplot [,domain=0:1,samples=2,style=semithick,color=green]({372.8075 + 31.5644*x},{517.5281 + 3.471*x});  
\addplot [,domain=0:1,samples=2,style=semithick,color=green]({292.2312 + 23.786*x},{509.6426 + 5.3486*x});  
\addplot [,domain=0:1,samples=2,style=semithick,color=green]({316.0172 + 28.3659*x},{514.9912 + 4.7857*x});  
\addplot [,domain=0:1,samples=2,style=semithick,color=green]({236.3359 + 23.6457*x},{507.4211 + 4.538*x});  
\addplot [,domain=0:1,samples=2,style=semithick,color=green]({259.9816 + 28.1759*x},{511.9591 + 4.1209*x});  
\addplot [,domain=0:1,samples=2,style=semithick,color=green]({182.9524 + 25.2085*x},{505.1413 + 5.1553*x});  
\addplot [,domain=0:1,samples=2,style=semithick,color=green]({208.1609 + 26.6232*x},{510.2966 + 5.3856*x});  
\addplot [,domain=0:1,samples=2,style=semithick,color=green]({127.6602 + 27.3743*x},{509.1443 + -0.82044*x});  
\addplot [,domain=0:1,samples=2,style=semithick,color=green]({155.0344 + 27.3289*x},{508.3239 + -0.31596*x});  
\addplot [,domain=0:1,samples=2,style=semithick,color=green]({86.1895 + 19.2388*x},{505.2321 + 1.3767*x});  
\addplot [,domain=0:1,samples=2,style=semithick,color=green]({105.4283 + 22.0704*x},{506.6088 + 1.5679*x});  
\addplot [,domain=0:1,samples=2,style=semithick,color=green]({83.0895 + 13.0211*x},{505.7489 + -10.009*x});  
\addplot [,domain=0:1,samples=2,style=semithick,color=green]({96.1106 + 14.5067*x},{495.74 + -9.9588*x});  
\addplot [,domain=0:1,samples=2,style=semithick,color=green]({115.7359 + 19.3262*x},{485.1783 + 4.1097*x});  
\addplot [,domain=0:1,samples=2,style=semithick,color=green]({135.0621 + 19.0071*x},{489.288 + 4.2644*x});  
\addplot [,domain=0:1,samples=2,style=semithick,color=green]({155.7839 + 15.496*x},{485.5047 + 10.0297*x});  
\addplot [,domain=0:1,samples=2,style=semithick,color=green]({171.2799 + 13.7276*x},{495.5343 + 10.0661*x});  
\addplot [,domain=0:1,samples=2,style=semithick,color=green]({182.1436 + 7.8647*x},{489.8463 + 0.93929*x});  
\addplot [,domain=0:1,samples=2,style=semithick,color=green]({190.0083 + 10.611*x},{490.7856 + 0.66835*x});  
\addplot [,domain=0:1,samples=2,style=semithick,color=green]({195.0905 + 8.2275*x},{487.6113 + 6.3878*x});  
\addplot [,domain=0:1,samples=2,style=semithick,color=green]({203.318 + 11.1855*x},{493.9991 + 6.1311*x});  
\addplot [,domain=0:1,samples=2,style=semithick,color=green]({214.6902 + 6.9783*x},{484.9489 + 6.7777*x});  
\addplot [,domain=0:1,samples=2,style=semithick,color=green]({221.6685 + 9.0259*x},{491.7265 + 7.3147*x});  
\addplot [,domain=0:1,samples=2,style=semithick,color=green]({229.5976 + 10.2744*x},{492.248 + 0.30305*x});  
\addplot [,domain=0:1,samples=2,style=semithick,color=green]({239.872 + 11.9114*x},{492.5511 + 0.22851*x});  
\addplot [,domain=0:1,samples=2,style=semithick,color=green]({249.3829 + 9.661*x},{490.1964 + 2.394*x});  
\addplot [,domain=0:1,samples=2,style=semithick,color=green]({259.0438 + 11.5161*x},{492.5904 + 2.6192*x});  
\addplot [,domain=0:1,samples=2,style=semithick,color=green]({266.536 + 10.7433*x},{490.203 + 2.0991*x});  
\addplot [,domain=0:1,samples=2,style=semithick,color=green]({277.2793 + 12.5811*x},{492.3021 + 2.4771*x});  
\addplot [,domain=0:1,samples=2,style=semithick,color=green]({284.8874 + 11.0997*x},{489.4234 + 2.6517*x});  
\addplot [,domain=0:1,samples=2,style=semithick,color=green]({295.9871 + 13.2894*x},{492.0751 + 2.93*x});  
\addplot [,domain=0:1,samples=2,style=semithick,color=green]({300.9813 + 10.7236*x},{489.3727 + 2.1697*x});  
\addplot [,domain=0:1,samples=2,style=semithick,color=green]({311.7049 + 13.4209*x},{491.5424 + 2.1494*x});  
\addplot [,domain=0:1,samples=2,style=semithick,color=green]({318.7431 + 9.8287*x},{490.0368 + 0.79768*x});  
\addplot [,domain=0:1,samples=2,style=semithick,color=green]({328.5718 + 12.5494*x},{490.8345 + 0.75361*x});  
\addplot [,domain=0:1,samples=2,style=semithick,color=green]({334.1273 + 9.9792*x},{489.2465 + 0.1331*x});  
\addplot [,domain=0:1,samples=2,style=semithick,color=green]({344.1065 + 12.585*x},{489.3796 + -0.10985*x});  
\addplot [,domain=0:1,samples=2,style=semithick,color=green]({351.2407 + 9.2488*x},{490.0279 + -0.72338*x});  
\addplot [,domain=0:1,samples=2,style=semithick,color=green]({360.4895 + 11.4412*x},{489.3045 + -1.2245*x});  
\addplot [,domain=0:1,samples=2,style=semithick,color=green]({366.1388 + 9.4763*x},{489.7616 + -1.1024*x});  
\addplot [,domain=0:1,samples=2,style=semithick,color=green]({375.6151 + 11.6852*x},{488.6591 + -1.3602*x});  
\end{axis} 
\end{tikzpicture} 

  \noindent}} &
\resizebox {!}{\sz\textwidth} {\resizebox {!}{0.2\textwidth} {\begin{tikzpicture} 
\begin{axis}[y dir=reverse, 
 xmin=1,xmax=960, 
 ymin=1,ymax=600, 
 xticklabels = \empty, yticklabels = \empty, 
 grid=none, axis equal image] 
\addplot graphics[xmin=1,xmax=960,ymin=1,ymax=600] {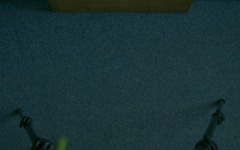}; 
\addplot [,domain=0:1,samples=2,style=semithick,color=green]({231.9942 + 18.3535*x},{555.3667 + 11.6587*x});  
\addplot [,domain=0:1,samples=2,style=semithick,color=green]({250.3477 + 0.55792*x},{567.0254 + 10.7162*x});  
\addplot [,domain=0:1,samples=2,style=semithick,color=green]({236.6898 + 25.5472*x},{557.1025 + -41.4492*x});  
\addplot [,domain=0:1,samples=2,style=semithick,color=green]({262.237 + 13.4358*x},{515.6533 + -44.6334*x});  
\addplot [,domain=0:1,samples=2,style=semithick,color=green]({275.7994 + 12.8037*x},{458.4539 + -32.9572*x});  
\addplot [,domain=0:1,samples=2,style=semithick,color=green]({288.6031 + 8.7472*x},{425.4967 + -36.5116*x});  
\addplot [,domain=0:1,samples=2,style=semithick,color=green]({303.4271 + 8.171*x},{357.7241 + -15.755*x});  
\addplot [,domain=0:1,samples=2,style=semithick,color=green]({311.5981 + 6.2134*x},{341.9691 + -21.1329*x});  
\addplot [,domain=0:1,samples=2,style=semithick,color=green]({323.9675 + 14.0535*x},{274.425 + -14.2722*x});  
\addplot [,domain=0:1,samples=2,style=semithick,color=green]({338.021 + 5.0619*x},{260.1528 + -18.7748*x});  
\addplot [,domain=0:1,samples=2,style=semithick,color=green]({339.0822 + 16.5519*x},{214.5387 + -30.3291*x});  
\addplot [,domain=0:1,samples=2,style=semithick,color=green]({355.6341 + 6.0705*x},{184.2096 + -33.7672*x});  
\addplot [,domain=0:1,samples=2,style=semithick,color=green]({359.7762 + 9.8055*x},{123.6292 + -20.2318*x});  
\addplot [,domain=0:1,samples=2,style=semithick,color=green]({369.5817 + 11.9376*x},{103.3973 + -20.8101*x});  
\addplot [,domain=0:1,samples=2,style=semithick,color=green]({378.2928 + 6.354*x},{61.279 + 17.2539*x});  
\addplot [,domain=0:1,samples=2,style=semithick,color=green]({384.6468 + 8.5198*x},{78.5329 + 17.8344*x});  
\addplot [,domain=0:1,samples=2,style=semithick,color=green]({387.6277 + 6.5518*x},{89.2666 + 17.5489*x});  
\addplot [,domain=0:1,samples=2,style=semithick,color=green]({394.1795 + 8.9376*x},{106.8156 + 17.072*x});  
\addplot [,domain=0:1,samples=2,style=semithick,color=green]({396.451 + 5.8168*x},{119.1417 + 10.8727*x});  
\addplot [,domain=0:1,samples=2,style=semithick,color=green]({402.2678 + 7.9232*x},{130.0144 + 10.5607*x});  
\addplot [,domain=0:1,samples=2,style=semithick,color=green]({404.425 + 4.5601*x},{141.0593 + 10.6491*x});  
\addplot [,domain=0:1,samples=2,style=semithick,color=green]({408.9851 + 6.916*x},{151.7084 + 10.7553*x});  
\addplot [,domain=0:1,samples=2,style=semithick,color=green]({408.775 + 4.4514*x},{164.6856 + 7.1979*x});  
\addplot [,domain=0:1,samples=2,style=semithick,color=green]({413.2264 + 7.684*x},{171.8835 + 7.1942*x});  
\addplot [,domain=0:1,samples=2,style=semithick,color=green]({413.0235 + 4.1949*x},{186.3844 + 4.2895*x});  
\addplot [,domain=0:1,samples=2,style=semithick,color=green]({417.2184 + 7.2512*x},{190.674 + 4.4402*x});  
\addplot [,domain=0:1,samples=2,style=semithick,color=green]({416.5407 + 4.2771*x},{207.3022 + 2.3847*x});  
\addplot [,domain=0:1,samples=2,style=semithick,color=green]({420.8178 + 7.0415*x},{209.6869 + 2.4019*x});  
\addplot [,domain=0:1,samples=2,style=semithick,color=green]({416.8853 + 5.7154*x},{222.7672 + 5.0625*x});  
\addplot [,domain=0:1,samples=2,style=semithick,color=green]({422.6007 + 7.6756*x},{227.8297 + 5.3065*x});  
\addplot [,domain=0:1,samples=2,style=semithick,color=green]({418.9933 + 6.5348*x},{240.2495 + 4.59*x});  
\addplot [,domain=0:1,samples=2,style=semithick,color=green]({425.5281 + 10.0398*x},{244.8395 + 4.1599*x});  
\end{axis} 
\end{tikzpicture} 

  \noindent}} &
\resizebox {!}{\sz\textwidth} {\resizebox {!}{0.2\textwidth} {\begin{tikzpicture} 
\begin{axis}[y dir=reverse, 
 xmin=1,xmax=1920, 
 ymin=1,ymax=1080, 
 xticklabels = \empty, yticklabels = \empty, 
 grid=none, axis equal image] 
\addplot graphics[xmin=1,xmax=1920,ymin=1,ymax=1080] {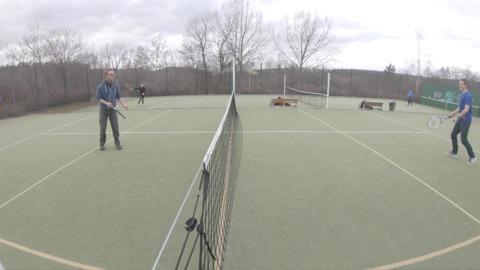}; 
\addplot [,domain=0:1,samples=2,style=semithick,color=green]({1601.9596 + 17.7669*x},{389.5026 + 28.571*x});  
\addplot [,domain=0:1,samples=2,style=semithick,color=green]({1619.7265 + 13.8876*x},{418.0735 + 31.4598*x});  
\addplot [,domain=0:1,samples=2,style=semithick,color=green]({1574.5348 + 11.1722*x},{369.3922 + 9.805*x});  
\addplot [,domain=0:1,samples=2,style=semithick,color=green]({1585.7069 + 12.7906*x},{379.1973 + 9.9967*x});  
\addplot [,domain=0:1,samples=2,style=semithick,color=green]({1551.8796 + 10.9104*x},{349.6176 + 9.9467*x});  
\addplot [,domain=0:1,samples=2,style=semithick,color=green]({1562.79 + 12.001*x},{359.5643 + 9.6531*x});  
\addplot [,domain=0:1,samples=2,style=semithick,color=green]({1527.0665 + 12.7206*x},{336.1712 + 7.4169*x});  
\addplot [,domain=0:1,samples=2,style=semithick,color=green]({1539.7872 + 12.8744*x},{343.5881 + 6.6545*x});  
\addplot [,domain=0:1,samples=2,style=semithick,color=green]({1503.7228 + 12.566*x},{318.7392 + 9.0691*x});  
\addplot [,domain=0:1,samples=2,style=semithick,color=green]({1516.2888 + 12.8557*x},{327.8082 + 9.2195*x});  
\addplot [,domain=0:1,samples=2,style=semithick,color=green]({1476.3037 + 13.4812*x},{308.4871 + 6.388*x});  
\addplot [,domain=0:1,samples=2,style=semithick,color=green]({1489.7849 + 13.8728*x},{314.875 + 6.0162*x});  
\addplot [,domain=0:1,samples=2,style=semithick,color=green]({1449.6156 + 13.1892*x},{298.6118 + 5.6435*x});  
\addplot [,domain=0:1,samples=2,style=semithick,color=green]({1462.8048 + 13.5171*x},{304.2554 + 5.517*x});  
\addplot [,domain=0:1,samples=2,style=semithick,color=green]({1420.291 + 14.2685*x},{290.6106 + 3.9147*x});  
\addplot [,domain=0:1,samples=2,style=semithick,color=green]({1434.5595 + 14.8306*x},{294.5254 + 4.0587*x});  
\addplot [,domain=0:1,samples=2,style=semithick,color=green]({1389.8435 + 14.9712*x},{284.1891 + 3.2911*x});  
\addplot [,domain=0:1,samples=2,style=semithick,color=green]({1404.8147 + 15.1301*x},{287.4802 + 3.3703*x});  
\addplot [,domain=0:1,samples=2,style=semithick,color=green]({1357.907 + 16.6252*x},{281.3781 + 1.6567*x});  
\addplot [,domain=0:1,samples=2,style=semithick,color=green]({1374.5322 + 16.3742*x},{283.0348 + 1.8738*x});  
\addplot [,domain=0:1,samples=2,style=semithick,color=green]({1327.3865 + 16.3016*x},{280.8045 + 0.41328*x});  
\addplot [,domain=0:1,samples=2,style=semithick,color=green]({1343.688 + 15.9618*x},{281.2178 + 0.56011*x});  
\addplot [,domain=0:1,samples=2,style=semithick,color=green]({1292.4495 + 17.4515*x},{280.9813 + 0.18312*x});  
\addplot [,domain=0:1,samples=2,style=semithick,color=green]({1309.901 + 17.0241*x},{281.1644 + 0.34495*x});  
\addplot [,domain=0:1,samples=2,style=semithick,color=green]({1258.1497 + 17.0645*x},{285.6407 + -2.2323*x});  
\addplot [,domain=0:1,samples=2,style=semithick,color=green]({1275.2142 + 18.0387*x},{283.4083 + -2.2416*x});  
\addplot [,domain=0:1,samples=2,style=semithick,color=green]({1222.6709 + 16.8768*x},{292.1141 + -3.396*x});  
\addplot [,domain=0:1,samples=2,style=semithick,color=green]({1239.5477 + 17.9202*x},{288.7181 + -3.4502*x});  
\addplot [,domain=0:1,samples=2,style=semithick,color=green]({1186.106 + 17.2725*x},{301.5378 + -3.9561*x});  
\addplot [,domain=0:1,samples=2,style=semithick,color=green]({1203.3784 + 18.2059*x},{297.5818 + -4.2622*x});  
\addplot [,domain=0:1,samples=2,style=semithick,color=green]({1148.0013 + 18.9977*x},{312.7886 + -4.9082*x});  
\addplot [,domain=0:1,samples=2,style=semithick,color=green]({1166.9991 + 19.0715*x},{307.8804 + -5.2147*x});  
\addplot [,domain=0:1,samples=2,style=semithick,color=green]({1110.9144 + 18.2927*x},{330.6309 + -7.9161*x});  
\addplot [,domain=0:1,samples=2,style=semithick,color=green]({1129.2071 + 18.7414*x},{322.7148 + -8.0741*x});  
\addplot [,domain=0:1,samples=2,style=semithick,color=green]({1071.6842 + 19.0512*x},{345.2567 + -7.8709*x});  
\addplot [,domain=0:1,samples=2,style=semithick,color=green]({1090.7354 + 19.6749*x},{337.3858 + -7.8096*x});  
\addplot [,domain=0:1,samples=2,style=semithick,color=green]({1032.8811 + 19.099*x},{369.6834 + -10.3066*x});  
\addplot [,domain=0:1,samples=2,style=semithick,color=green]({1051.9801 + 18.8303*x},{359.3768 + -10.3753*x});  
\addplot [,domain=0:1,samples=2,style=semithick,color=green]({994.9501 + 18.4106*x},{394.6173 + -12.964*x});  
\addplot [,domain=0:1,samples=2,style=semithick,color=green]({1013.3607 + 18.6656*x},{381.6533 + -12.8064*x});  
\addplot [,domain=0:1,samples=2,style=semithick,color=green]({953.8009 + 19.6775*x},{421.4843 + -13.6334*x});  
\addplot [,domain=0:1,samples=2,style=semithick,color=green]({973.4784 + 20.2993*x},{407.8509 + -14.181*x});  
\addplot [,domain=0:1,samples=2,style=semithick,color=green]({917.9457 + 17.8172*x},{451.0426 + -15.899*x});  
\addplot [,domain=0:1,samples=2,style=semithick,color=green]({935.763 + 19.1279*x},{435.1437 + -15.7307*x});  
\addplot [,domain=0:1,samples=2,style=semithick,color=green]({880.2356 + 5.9501*x},{473.9074 + -3.6971*x});  
\addplot [,domain=0:1,samples=2,style=semithick,color=green]({886.1857 + 5.7215*x},{470.2103 + -2.8511*x});  
\addplot [,domain=0:1,samples=2,style=semithick,color=green]({844.1927 + 13.9544*x},{515.2476 + -16.1993*x});  
\addplot [,domain=0:1,samples=2,style=semithick,color=green]({858.1471 + 15.1507*x},{499.0483 + -14.9722*x});  
\addplot [,domain=0:1,samples=2,style=semithick,color=green]({810.6716 + 16.4372*x},{545.2408 + -16.1053*x});  
\addplot [,domain=0:1,samples=2,style=semithick,color=green]({827.1088 + 17.0704*x},{529.1355 + -16.4425*x});  
\addplot [,domain=0:1,samples=2,style=semithick,color=green]({777.254 + 3.1617*x},{580.6295 + -2.5594*x});  
\addplot [,domain=0:1,samples=2,style=semithick,color=green]({780.4156 + 4.0004*x},{578.0701 + -2.3747*x});  
\addplot [,domain=0:1,samples=2,style=semithick,color=green]({740.8246 + 5.4388*x},{619.4097 + -1.4149*x});  
\addplot [,domain=0:1,samples=2,style=semithick,color=green]({746.2633 + 5.6797*x},{617.9948 + -5.0375*x});  
\addplot [,domain=0:1,samples=2,style=semithick,color=green]({727.7373 + 8.237*x},{604.9638 + 10.7612*x});  
\addplot [,domain=0:1,samples=2,style=semithick,color=green]({735.9743 + 8.0569*x},{615.725 + 10.1722*x});  
\addplot [,domain=0:1,samples=2,style=semithick,color=green]({712.8297 + 7.1167*x},{577.8051 + 15.3953*x});  
\addplot [,domain=0:1,samples=2,style=semithick,color=green]({719.9464 + 8.5153*x},{593.2003 + 15.4828*x});  
\addplot [,domain=0:1,samples=2,style=semithick,color=green]({694.4211 + 8.0222*x},{554.1018 + 10.2303*x});  
\addplot [,domain=0:1,samples=2,style=semithick,color=green]({702.4433 + 7.6681*x},{564.3321 + 9.904*x});  
\addplot [,domain=0:1,samples=2,style=semithick,color=green]({679.9833 + 8.8782*x},{531.3231 + 10.7701*x});  
\addplot [,domain=0:1,samples=2,style=semithick,color=green]({688.8616 + 8.7728*x},{542.0932 + 10.8663*x});  
\addplot [,domain=0:1,samples=2,style=semithick,color=green]({662.5179 + 10.0553*x},{511.3991 + 10.5153*x});  
\addplot [,domain=0:1,samples=2,style=semithick,color=green]({672.5732 + 10.4604*x},{521.9145 + 10.1625*x});  
\addplot [,domain=0:1,samples=2,style=semithick,color=green]({648.3313 + 5.5069*x},{502.8051 + 3.8017*x});  
\addplot [,domain=0:1,samples=2,style=semithick,color=green]({653.8382 + 6.5127*x},{506.6068 + 3.3182*x});  
\addplot [,domain=0:1,samples=2,style=semithick,color=green]({633.1551 + 7.9714*x},{481.4738 + 11.9353*x});  
\addplot [,domain=0:1,samples=2,style=semithick,color=green]({641.1265 + 9.6475*x},{493.4091 + 12.0963*x});  
\addplot [,domain=0:1,samples=2,style=semithick,color=green]({614.7123 + 7.9289*x},{463.8869 + 13.7007*x});  
\addplot [,domain=0:1,samples=2,style=semithick,color=green]({622.6412 + 10.1065*x},{477.5877 + 13.9279*x});  
\addplot [,domain=0:1,samples=2,style=semithick,color=green]({604.2987 + 7.9122*x},{455.8039 + 12.1458*x});  
\addplot [,domain=0:1,samples=2,style=semithick,color=green]({612.2109 + 11.8576*x},{467.9497 + 12.2593*x});  
\end{axis} 
\end{tikzpicture} 

  \noindent}} &
\resizebox {!}{\sz\textwidth} {\input{imgs/deepfmo/throw_tennis}} \\

\rotatebox{90}{\hspace{10pt} + deblurring} &
\resizebox {!}{\sz\textwidth} {\resizebox {!}{0.2\textwidth} {\begin{tikzpicture} 
\begin{axis}[y dir=reverse, 
 xmin=1,xmax=960, 
 ymin=1,ymax=600, 
 xticklabels = \empty, yticklabels = \empty, 
 grid=none, axis equal image] 
\addplot graphics[xmin=1,xmax=960,ymin=1,ymax=600] {imgs/thumbnails/tbd/hit_tennis.jpg};

\addplot [,domain=0:1,samples=2,style=semithick,color=green]({701.9524 + 70.5151*x},{516.2025 + -2.5867*x});  
\addplot [,domain=0:1,samples=2,style=semithick,color=green]({647.9803 + 59.9883*x},{521.908 + 0.83742*x});  
\addplot [,domain=0:1,samples=2,style=semithick,color=green]({588.0393 + 58.5374*x},{520.6752 + 1.7366*x});  
\addplot [,domain=0:1,samples=2,style=semithick,color=green]({528.0204 + 58.5798*x},{520.0852 + 2.8733*x});  
\addplot [,domain=0:1,samples=2,style=semithick,color=green]({466.4793 + 58.5481*x},{520.6111 + 0.75094*x});  
\addplot [,domain=0:1,samples=10,style=semithick,color=green]({407.859 + 62.3959*x + -4.9618*x^2},{519.8342 + -4.4095*x + 6.0658*x^2});  
\addplot [,domain=0:1,samples=2,style=semithick,color=green]({349.3901 + 58.5092*x},{516.7892 + 2.8085*x});  
\addplot [,domain=0:1,samples=10,style=semithick,color=green]({293.7761 + 26.6575*x + 30.255*x^2},{515.5694 + -4.1378*x + 7.2846*x^2});  
\addplot [,domain=0:1,samples=10,style=semithick,color=green]({239.6083 + 33.0788*x + 20.9901*x^2},{515.4032 + -7.0276*x + 8.5983*x^2});  
\addplot [,domain=0:1,samples=2,style=semithick,color=green]({187.4463 + 53.974*x},{511.3464 + 2.1522*x});  
\addplot [,domain=0:1,samples=10,style=semithick,color=green]({134.9565 + 57.9612*x + -6.5484*x^2},{511.0451 + -5.6014*x + 6.9861*x^2});  
\addplot [,domain=0:1,samples=10,style=semithick,color=green]({91.4213 + 40.4164*x + 2.0174*x^2},{507.5914 + -1.5362*x + 5.0644*x^2});  
\addplot [,domain=0:1,samples=2,style=semithick,color=green]({85.1091 + 29.2338*x},{501.3352 + -9.8136*x});  
\addplot [,domain=0:1,samples=10,style=semithick,color=green]({118.5905 + 46.4144*x + -5.2894*x^2},{493.8505 + -9.857*x + 15.7324*x^2});  
\addplot [,domain=0:1,samples=2,style=semithick,color=green]({159.0167 + 28.507*x},{497.893 + 4.455*x});  
\addplot [,domain=0:1,samples=2,style=semithick,color=green]({185.6142 + 13.6304*x},{497.0947 + -2.186*x});  
\addplot [,domain=0:1,samples=2,style=semithick,color=green]({199.8781 + 15.5142*x},{495.7671 + 8.0032*x});  
\addplot [,domain=0:1,samples=10,style=semithick,color=green]({215.7357 + 20.1401*x + 0.42282*x^2},{500.3452 + -11.818*x + 9.9835*x^2});  
\addplot [,domain=0:1,samples=2,style=semithick,color=green]({235.2706 + 19.7351*x},{500.9389 + -2.9001*x});  
\addplot [,domain=0:1,samples=2,style=semithick,color=green]({254.9859 + 18.1226*x},{498.1542 + 1.6629*x});  
\addplot [,domain=0:1,samples=10,style=semithick,color=green]({272.6156 + 18.2124*x + 0.10061*x^2},{498.9278 + -4.1883*x + 4.8034*x^2});  
\addplot [,domain=0:1,samples=2,style=semithick,color=green]({290.959 + 16.5656*x},{497.8907 + -0.42171*x});  
\addplot [,domain=0:1,samples=2,style=semithick,color=green]({307.4527 + 16.6328*x},{497.393 + -1.2962*x});  
\addplot [,domain=0:1,samples=2,style=semithick,color=green]({325.2768 + 15.3276*x},{496.3274 + -2.0458*x});  
\addplot [,domain=0:1,samples=2,style=semithick,color=green]({341.8268 + 15.1473*x},{494.8952 + -1.6345*x});  
\addplot [,domain=0:1,samples=2,style=semithick,color=green]({356.9429 + 15.1001*x},{493.654 + -0.64029*x});  
\addplot [,domain=0:1,samples=2,style=semithick,color=green]({371.9557 + 16.533*x},{492.6641 + -0.87629*x});  
\end{axis} 
\end{tikzpicture} 

  \noindent}} &
\resizebox {!}{\sz\textwidth} {\resizebox {!}{0.2\textwidth} {\begin{tikzpicture} 
\begin{axis}[y dir=reverse, 
 xmin=1,xmax=960, 
 ymin=1,ymax=600, 
 xticklabels = \empty, yticklabels = \empty, 
 grid=none, axis equal image] 
\addplot graphics[xmin=1,xmax=960,ymin=1,ymax=600] {imgs/thumbnails/tbd/roll_golf.jpg}; 

\addplot [,domain=0:1,samples=2,style=semithick,color=green]({240.0007 + 14.9991*x},{580.5926 + -0.11703*x});  
\addplot [,domain=0:1,samples=10,style=semithick,color=green]({281.4917 + -13.6677*x + -14.5794*x^2},{462.5613 + 63.1907*x + 30.6671*x^2});  
\addplot [,domain=0:1,samples=2,style=semithick,color=green]({278.3731 + 26.5038*x},{467.8094 + -87.1509*x});  
\addplot [,domain=0:1,samples=2,style=semithick,color=green]({303.2684 + 25.0843*x},{381.08 + -84.124*x});  
\addplot [,domain=0:1,samples=10,style=semithick,color=green]({348.5857 + -17.102*x + -6.7571*x^2},{218.028 + 67.7038*x + 14.0198*x^2});  
\addplot [,domain=0:1,samples=10,style=semithick,color=green]({365.9484 + -10.7007*x + -10.0989*x^2},{141.2573 + 59.8293*x + 18.984*x^2});  
\addplot [,domain=0:1,samples=2,style=semithick,color=green]({363.8404 + 18.6184*x},{143.4124 + -68.4236*x});  
\addplot [,domain=0:1,samples=10,style=semithick,color=green]({384.0214 + 12.2805*x + -6.2155*x^2},{64.5177 + 25.2917*x + 1.6688*x^2});  
\addplot [,domain=0:1,samples=2,style=semithick,color=green]({401.5459 + -11.2368*x},{118.0331 + -25.1711*x});  
\addplot [,domain=0:1,samples=10,style=semithick,color=green]({398.5831 + 15.0989*x + -6.1134*x^2},{118.8829 + 18.9778*x + 1.6579*x^2});  
\addplot [,domain=0:1,samples=10,style=semithick,color=green]({404.4759 + 18.4883*x + -10.4398*x^2},{140.0724 + 20.6049*x + 1.3639*x^2});  
\addplot [,domain=0:1,samples=2,style=semithick,color=green]({417.6928 + -3.5352*x},{182.6005 + -19.4023*x});  
\addplot [,domain=0:1,samples=2,style=semithick,color=green]({420.2248 + -3.1086*x},{200.7018 + -17.981*x});  
\addplot [,domain=0:1,samples=2,style=semithick,color=green]({422.1992 + -2.1203*x},{220.2398 + -17.9265*x});  
\addplot [,domain=0:1,samples=10,style=semithick,color=green]({425.0233 + -1.7462*x + -2.0537*x^2},{236.723 + -17.0553*x + 0.46236*x^2});  
\addplot [,domain=0:1,samples=2,style=semithick,color=green]({430.1147 + -5.7818*x},{251.5259 + -13.5928*x});  
 
\end{axis} 
\end{tikzpicture} 

  \noindent}} &
\resizebox {!}{\sz\textwidth} {\resizebox {!}{0.2\textwidth} {\begin{tikzpicture} 
\begin{axis}[y dir=reverse, 
 xmin=1,xmax=1920, 
 ymin=1,ymax=1080, 
 xticklabels = \empty, yticklabels = \empty, 
 grid=none, axis equal image] 
\addplot graphics[xmin=1,xmax=1920,ymin=1,ymax=1080] {imgs/thumbnails/tbd/tennis.jpg}; 
\addplot [,domain=0:1,samples=2,style=semithick,color=green]({1619.2224 + -21.0942*x},{414.6372 + -25.7422*x});  
\addplot [,domain=0:1,samples=2,style=semithick,color=green]({1577.8362 + 19.5416*x},{372.1828 + 17.5115*x});  
\addplot [,domain=0:1,samples=2,style=semithick,color=green]({1554.8461 + 20.8452*x},{353.1866 + 17.1877*x});  
\addplot [,domain=0:1,samples=2,style=semithick,color=green]({1531.2378 + 21.7845*x},{337.6615 + 15.3067*x});  
\addplot [,domain=0:1,samples=2,style=semithick,color=green]({1505.8363 + 23.6444*x},{322.2584 + 14.9831*x});  
\addplot [,domain=0:1,samples=2,style=semithick,color=green]({1479.0762 + 26.0957*x},{310.8314 + 11.7882*x});  
\addplot [,domain=0:1,samples=2,style=semithick,color=green]({1452.226 + 26.1154*x},{300.4494 + 10.7189*x});  
\addplot [,domain=0:1,samples=2,style=semithick,color=green]({1423.1788 + 27.9248*x},{292.3952 + 8.2544*x});  
\addplot [,domain=0:1,samples=2,style=semithick,color=green]({1393.9091 + 28.0689*x},{286.451 + 5.6579*x});  
\addplot [,domain=0:1,samples=2,style=semithick,color=green]({1361.9613 + 30.0989*x},{283.387 + 3.0104*x});  
\addplot [,domain=0:1,samples=10,style=semithick,color=green]({1330.9411 + 28.3725*x + 1.7762*x^2},{283.2126 + -2.3209*x + 3.468*x^2});  
\addplot [,domain=0:1,samples=2,style=semithick,color=green]({1296.0023 + 33.0063*x},{282.7454 + 0.29934*x});  
\addplot [,domain=0:1,samples=2,style=semithick,color=green]({1260.9769 + 33.0714*x},{286.7671 + -3.2805*x});  
\addplot [,domain=0:1,samples=2,style=semithick,color=green]({1224.9741 + 34.0294*x},{293.8715 + -6.8542*x});  
\addplot [,domain=0:1,samples=2,style=semithick,color=green]({1188.0796 + 35.9295*x},{303.3085 + -9.2732*x});  
\addplot [,domain=0:1,samples=2,style=semithick,color=green]({1150.047 + 36.958*x},{315.1559 + -11.1394*x});  
\addplot [,domain=0:1,samples=2,style=semithick,color=green]({1112.9161 + 38.7551*x},{330.778 + -14.648*x});  
\addplot [,domain=0:1,samples=2,style=semithick,color=green]({1075.0988 + 37.5319*x},{349.2074 + -17.8836*x});  
\addplot [,domain=0:1,samples=2,style=semithick,color=green]({1034.7609 + 38.3557*x},{371.5706 + -21.3612*x});  
\addplot [,domain=0:1,samples=2,style=semithick,color=green]({997.9762 + 36.8377*x},{394.9639 + -24.2465*x});  
\addplot [,domain=0:1,samples=2,style=semithick,color=green]({957.1153 + 38.0519*x},{422.1693 + -25.9238*x});  
\addplot [,domain=0:1,samples=2,style=semithick,color=green]({920.7836 + 35.2304*x},{450.7345 + -28.7173*x});  
\addplot [,domain=0:1,samples=2,style=semithick,color=green]({923.0517 + -3.0242*x},{466.0956 + -29.9975*x});  
\addplot [,domain=0:1,samples=10,style=semithick,color=green]({867.8524 + -24.3673*x + 5.803*x^2},{491.801 + 15.4774*x + 3.4666*x^2});  
\addplot [,domain=0:1,samples=2,style=semithick,color=green]({814.4133 + 31.0511*x},{545.3916 + -29.947*x});  
\addplot [,domain=0:1,samples=2,style=semithick,color=green]({788 + 1*x},{576 + 1*x});  
\addplot [,domain=0:1,samples=2,style=semithick,color=green]({744.9768 + -0.012039*x},{632 + -18*x});  
\addplot [,domain=0:1,samples=2,style=semithick,color=green]({735.1915 + -5.5016*x},{613.8786 + -8.6822*x});  
\addplot [,domain=0:1,samples=10,style=semithick,color=green]({715.685 + 7.9452*x + 4.714*x^2},{577.7876 + 24.4601*x + -2.5721*x^2});  
\addplot [,domain=0:1,samples=2,style=semithick,color=green]({712.8173 + -14.4322*x},{576.4477 + -21.3564*x});  
\addplot [,domain=0:1,samples=2,style=semithick,color=green]({694.8136 + -11.4671*x},{554.1166 + -18.3333*x});  
\addplot [,domain=0:1,samples=2,style=semithick,color=green]({682.9195 + -15.9997*x},{533.0711 + -18.117*x});  
\addplot [,domain=0:1,samples=10,style=semithick,color=green]({651.5645 + 11.1701*x + -2.356*x^2},{499.4938 + 11.9108*x + 1.4135*x^2});  
\addplot [,domain=0:1,samples=2,style=semithick,color=green]({644.6582 + -7.0511*x},{492.275 + -8.7635*x});  
\addplot [,domain=0:1,samples=2,style=semithick,color=green]({570 + -1.7764e-15*x},{420 + 29*x});  
\end{axis} 
\end{tikzpicture} 

  \noindent}} & 
\resizebox {!}{\sz\textwidth} {\resizebox {!}{0.2\textwidth} {\begin{tikzpicture} 
\begin{axis}[y dir=reverse, 
 xmin=1,xmax=960, 
 ymin=1,ymax=600, 
 xticklabels = \empty, yticklabels = \empty, 
 grid=none, axis equal image] 
\addplot graphics[xmin=1,xmax=960,ymin=1,ymax=600] {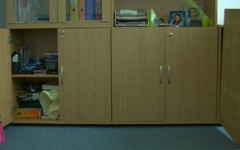}; 
\addplot [,domain=0:1,samples=2,style=semithick,color=green]({846.3513 + -72.3615*x},{94.1723 + -77.5622*x});  
\addplot [,domain=0:1,samples=2,style=semithick,color=green]({884.4327 + 2.3458*x},{195.0486 + -45.0341*x});  
\addplot [,domain=0:1,samples=2,style=semithick,color=green]({768.0152 + 84.0015*x},{295.5154 + -82.4985*x});  
\addplot [,domain=0:1,samples=2,style=semithick,color=green]({676.7798 + 87.5384*x},{400.948 + -101.5358*x});  
\addplot [,domain=0:1,samples=2,style=semithick,color=green]({602.0072 + 72.3352*x},{503.2645 + -97.6404*x});  
\addplot [,domain=0:1,samples=2,style=semithick,color=green]({597.7099 + -74.8116*x},{515.4109 + -90.1563*x});  
\addplot [,domain=0:1,samples=2,style=semithick,color=green]({448.0319 + 69.9885*x},{348.4384 + 68.0263*x});  
\addplot [,domain=0:1,samples=2,style=semithick,color=green]({371.5959 + 70.225*x},{289.5442 + 52.8652*x});  
\addplot [,domain=0:1,samples=2,style=semithick,color=green]({295.9854 + 68.9759*x},{254.0711 + 36.0461*x});  
\addplot [,domain=0:1,samples=10,style=semithick,color=green]({215.6226 + 32.4445*x + -0.80824*x^2},{237.7623 + -6.2523*x + 9.0859*x^2});  
\addplot [,domain=0:1,samples=10,style=semithick,color=green]({143.5632 + 45.1733*x + 0.235*x^2},{231.4376 + -8.0187*x + 7.7396*x^2});  
\addplot [,domain=0:1,samples=10,style=semithick,color=green]({74.84 + 75.2988*x + -11.8319*x^2},{242.576 + -29.4411*x + 18.6478*x^2});  
\addplot [,domain=0:1,samples=10,style=semithick,color=green]({35.9543 + -26.9256*x + 55.9767*x^2},{274.0476 + -32.0181*x + 5.3334*x^2});  
\addplot [,domain=0:1,samples=2,style=semithick,color=green]({66.0571 + 51.4259*x},{256.3048 + 15.0441*x});  
\addplot [,domain=0:1,samples=2,style=semithick,color=green]({120.9475 + 52.3439*x},{272.8617 + 30.2698*x});  
\addplot [,domain=0:1,samples=2,style=semithick,color=green]({176.2828 + 64.2833*x},{302.3542 + 53.9716*x});  
\addplot [,domain=0:1,samples=2,style=semithick,color=green]({291.0943 + -53.0049*x},{416.9198 + -62.3459*x});  
\addplot [,domain=0:1,samples=2,style=semithick,color=green]({348.309 + -13.9029*x},{497.5591 + -12.9584*x});  
\addplot [,domain=0:1,samples=2,style=semithick,color=green]({354.3149 + 20.9992*x},{519.1377 + -48.0004*x});  
\addplot [,domain=0:1,samples=2,style=semithick,color=green]({379.6186 + 26.227*x},{467.1585 + -43.0572*x});  
\addplot [,domain=0:1,samples=10,style=semithick,color=green]({405.8357 + 20.8228*x + 4.3297*x^2},{417.9725 + -28.4114*x + 4.1655*x^2});  
\addplot [,domain=0:1,samples=2,style=semithick,color=green]({433.0445 + 23.2957*x},{388.7144 + -8.2542*x});  
\addplot [,domain=0:1,samples=10,style=semithick,color=green]({463.4822 + 25.7079*x + -1.212*x^2},{379.5189 + -1.6099*x + 6.4319*x^2});  
\addplot [,domain=0:1,samples=2,style=semithick,color=green]({495.3268 + 24.948*x},{380.552 + 18.2005*x});  
\addplot [,domain=0:1,samples=2,style=semithick,color=green]({548.7015 + -29.3823*x},{443.1141 + -39.8494*x});  
\addplot [,domain=0:1,samples=2,style=semithick,color=green]({575.5504 + -29.0197*x},{502.733 + -56.0079*x});  
\addplot [,domain=0:1,samples=2,style=semithick,color=green]({581.9188 + 23.8853*x},{526.7449 + -30.0909*x});  
\addplot [,domain=0:1,samples=2,style=semithick,color=green]({601.6968 + 28.7368*x},{495.4631 + -36.1181*x});  
\addplot [,domain=0:1,samples=2,style=semithick,color=green]({631.0247 + 25.989*x},{458.2845 + -17.2639*x});  
\addplot [,domain=0:1,samples=2,style=semithick,color=green]({661.4054 + 25.4353*x},{438.4237 + -2.236*x});  
\addplot [,domain=0:1,samples=10,style=semithick,color=green]({690.1096 + 27.5418*x + -2.4233*x^2},{439.2098 + 7.7117*x + 4.8924*x^2});  
\addplot [,domain=0:1,samples=10,style=semithick,color=green]({746.2195 + -23.4762*x + -5.0111*x^2},{487.6697 + -39.7755*x + 4.435*x^2});  
\addplot [,domain=0:1,samples=2,style=semithick,color=green]({771.8695 + -26.2791*x},{531.9689 + -43.0241*x});  
\addplot [,domain=0:1,samples=10,style=semithick,color=green]({768.6517 + 21.6482*x + 6.9372*x^2},{518.4572 + -34.9671*x + 4.4882*x^2});  
\addplot [,domain=0:1,samples=2,style=semithick,color=green]({799.1695 + 27.2917*x},{489.9363 + -16.0025*x});  
\addplot [,domain=0:1,samples=2,style=semithick,color=green]({827.9576 + 25.4744*x},{474.1566 + -0.80598*x});  
\addplot [,domain=0:1,samples=2,style=semithick,color=green]({866.9996 + -10.2945*x},{487.5005 + -9.2292*x});  
\addplot [,domain=0:1,samples=10,style=semithick,color=green]({881.371 + 27.3569*x + -20.4697*x^2},{495.762 + 23.8217*x + 1.7121*x^2});  
\addplot [,domain=0:1,samples=2,style=semithick,color=green]({884.5585 + -14.9929*x},{531.401 + -16.5064*x});  
\addplot [,domain=0:1,samples=2,style=semithick,color=green]({855.0257 + 15.3242*x},{512.7973 + 1.9439*x});  
\addplot [,domain=0:1,samples=2,style=semithick,color=green]({837.9556 + 12.4521*x},{526.0294 + -14.4058*x});  
\addplot [,domain=0:1,samples=2,style=semithick,color=green]({839.2557 + -13.8163*x},{530.2071 + -13.1681*x});  
\addplot [,domain=0:1,samples=2,style=semithick,color=green]({807.9401 + 13.021*x},{525.7443 + -9.6465*x});  
\addplot [,domain=0:1,samples=2,style=semithick,color=green]({794.1658 + 10.955*x},{523.5426 + 5.0034*x});  
\addplot [,domain=0:1,samples=2,style=semithick,color=green]({777.0496 + 13.7768*x},{529.5983 + -6.9515*x});  
\end{axis} 
\end{tikzpicture} 

  \noindent}} \\

\rotatebox{90}{\hspace{10pt}  + non-causal} &
\resizebox {!}{\sz\textwidth} {\resizebox {!}{0.2\textwidth} {\begin{tikzpicture} 
\begin{axis}[y dir=reverse, 
 xmin=1,xmax=960, 
 ymin=1,ymax=600, 
 xticklabels = \empty, yticklabels = \empty, 
 grid=none, axis equal image] 
\addplot graphics[xmin=1,xmax=960,ymin=1,ymax=600] {imgs/thumbnails/tbd/hit_tennis.jpg}; 
\addplot [>={Latex[length=1.5mm,width=0.5mm,angle'=25,open,round]},,domain=3:4,samples=5,style=semithick,color=green]({1073.5411 + -88.8292*x + 4.5178*x^2 + -0.31747*x^3 + 0.0088953*x^4},{446.0578 + 30.1773*x + -4.2319*x^2 + 0.24945*x^3 + -0.0054881*x^4});  
\addplot [>={Latex[length=1.5mm,width=0.5mm,angle'=25,open,round]},,domain=4:16,samples=120,style=semithick,color=green]({1073.5411 + -88.8292*x + 4.5178*x^2 + -0.31747*x^3 + 0.0088953*x^4},{446.0578 + 30.1773*x + -4.2319*x^2 + 0.24945*x^3 + -0.0054881*x^4});  
\addplot [>={Latex[length=1.5mm,width=0.5mm,angle'=25,open,round]},,domain=0:1,samples=2,style=semithick,color=green]({91.4213 + -6.3122*x},{507.5914 + -6.2561*x});  
\addplot [>={Latex[length=1.5mm,width=0.5mm,angle'=25,open,round]},,domain=16:18,samples=20,style=semithick,color=green]({1200.5708 + -164.8449*x + 5.9455*x^2},{2773.3751 + -267.5117*x + 7.8443*x^2});  
\addplot [>={Latex[length=1.5mm,width=0.5mm,angle'=25,open,round]},,domain=0:1,samples=2,style=semithick,color=green]({159.7156 + 21.7844*x},{499.7259 + 1.2741*x});  
\addplot [>={Latex[length=1.5mm,width=0.5mm,angle'=25,open,round]},,domain=0:1,samples=2,style=semithick,color=green]({181.5 + 4.1142*x},{501 + -3.9053*x});  
\addplot [>={Latex[length=1.5mm,width=0.5mm,angle'=25,open,round]},,domain=19:20,samples=10,style=semithick,color=green]({-73.3637 + 13.6304*x},{538.6285 + -2.186*x});  
\addplot [>={Latex[length=1.5mm,width=0.5mm,angle'=25,open,round]},,domain=0:1,samples=2,style=semithick,color=green]({199.2447 + 13.7553*x},{494.9087 + 7.5913*x});  
\addplot [>={Latex[length=1.5mm,width=0.5mm,angle'=25,open,round]},,domain=0:1,samples=2,style=semithick,color=green]({213 + 2.7357*x},{502.5 + -2.1548*x});  
\addplot [->,>={Latex[length=1.5mm,width=0.5mm,angle'=25,open,round]},,domain=21:32,samples=100,style=semithick,color=green]({-587.574 + 63.6175*x + -1.5676*x^2 + 0.017134*x^3},{415.3509 + 10.3208*x + -0.39396*x^2 + 0.0045345*x^3});  
\end{axis} 
\end{tikzpicture} 

  \noindent}} &
\resizebox {!}{\sz\textwidth} {\resizebox {!}{0.2\textwidth} {\begin{tikzpicture} 
\begin{axis}[y dir=reverse, 
 xmin=1,xmax=960, 
 ymin=1,ymax=600, 
 xticklabels = \empty, yticklabels = \empty, 
 grid=none, axis equal image] 
\addplot graphics[xmin=1,xmax=960,ymin=1,ymax=600] {imgs/thumbnails/tbd/roll_golf.jpg}; 
\addplot [>={Latex[length=1.5mm,width=0.5mm,angle'=25,open,round]},,domain=1:8,samples=70,style=semithick,color=green]({217.927 + 22.289*x + -0.21532*x^2},{639.944 + -57.7417*x + -1.6097*x^2});  
\addplot [>={Latex[length=1.5mm,width=0.5mm,angle'=25,open,round]},,domain=0:1,samples=2,style=semithick,color=green]({382.4587 + 1.5627*x},{74.9888 + -10.4711*x});  
\addplot [->,>={Latex[length=1.5mm,width=0.5mm,angle'=25,open,round]},,domain=8:17,samples=90,style=semithick,color=green]({231.1597 + 30.9823*x + -1.7957*x^2 + 0.038917*x^3},{-235.2675 + 47.9233*x + -1.4588*x^2 + 0.019073*x^3});  
\end{axis} 
\end{tikzpicture} 

  \noindent}} &
\resizebox {!}{\sz\textwidth} {\resizebox {0.33\textwidth}{!} {\begin{tikzpicture} 
\begin{axis}[y dir=reverse, 
 xmin=1,xmax=1920, 
 ymin=1,ymax=1080, 
 xticklabels = \empty, yticklabels = \empty, 
 grid=none, axis equal image] 
\addplot graphics[xmin=1,xmax=1920,ymin=1,ymax=1080] {imgs/thumbnails/tbd/tennis.jpg}; 
\addplot [>={Latex[length=1.5mm,width=0.5mm,angle'=25,open,round]},,domain=1:27,samples=260,style=semithick,color=green]({1642.2514 + -24.0635*x + 1.2801*x^2 + -0.26249*x^3 + 0.017393*x^4 + -0.00055239*x^5 + 6.9992e-06*x^6},{437.6525 + -23.1509*x + -0.032079*x^2 + 0.18162*x^3 + -0.014553*x^4 + 0.00056172*x^5 + -8.0379e-06*x^6});  
\addplot [>={Latex[length=1.5mm,width=0.5mm,angle'=25,open,round]},,domain=0:1,samples=2,style=semithick,color=green]({788 + -47*x},{576 + 40*x});  
\addplot [>={Latex[length=1.5mm,width=0.5mm,angle'=25,open,round]},,domain=0:1,samples=2,style=semithick,color=green]({741 + -5.8085*x},{616 + -2.1214*x});  
\addplot [->,>={Latex[length=1.5mm,width=0.5mm,angle'=25,open,round]},,domain=28:36,samples=50,style=semithick,color=green]({334.1321 + 38.0624*x + -0.84782*x^2},{708.1869 + 10.5612*x + -0.49748*x^2});   
\end{axis} 
\end{tikzpicture} 

  \noindent}} & 
\resizebox {!}{\sz\textwidth} {\resizebox {!}{0.2\textwidth} {\begin{tikzpicture} 
\begin{axis}[y dir=reverse, 
 xmin=1,xmax=960, 
 ymin=1,ymax=600, 
 xticklabels = \empty, yticklabels = \empty, 
 grid=none, axis equal image] 
\addplot graphics[xmin=1,xmax=960,ymin=1,ymax=600] {imgs/thumbnails/tbd/throw_tennis.jpg}; 

\addplot [>={Latex[length=1.5mm,width=0.5mm,angle'=25,open,round]},,domain=0:1,samples=2,style=semithick,color=green]({773.9898 + 110*x},{16.6102 + 130*x});  
\addplot [>={Latex[length=1.5mm,width=0.5mm,angle'=25,open,round]},,domain=0:1,samples=2,style=semithick,color=green]({883.9898 +  -281.9810*x},{146.6102 + 356.6547*x});  

\addplot [>={Latex[length=1.5mm,width=0.5mm,angle'=25,open,round]},,domain=0:1,samples=2,style=semithick,color=green]({602.0072 + -6.5072*x},{503.2645 + 6.7355*x});  
\addplot [>={Latex[length=1.5mm,width=0.5mm,angle'=25,open,round]},,domain=0:1,samples=2,style=semithick,color=green]({595.5 + -77.4796*x},{510 + -93.5354*x});  
\addplot [>={Latex[length=1.5mm,width=0.5mm,angle'=25,open,round]},,domain=7:14,samples=70,style=semithick,color=green]({1416.1677 + -160.1019*x + 4.5422*x^2},{1280.1102 + -172.9882*x + 7.0872*x^2});  
\addplot [>={Latex[length=1.5mm,width=0.5mm,angle'=25,open,round]},,domain=0:1,samples=2,style=semithick,color=green]({65.0054 + 1.0517*x},{247.3629 + 8.9419*x});  
\addplot [>={Latex[length=1.5mm,width=0.5mm,angle'=25,open,round]},,domain=14:18,samples=40,style=semithick,color=green]({-640.1939 + 45.9254*x + 0.32293*x^2},{1713.7575 + -216.304*x + 8.0143*x^2});  
\addplot [>={Latex[length=1.5mm,width=0.5mm,angle'=25,open,round]},,domain=0:1,samples=2,style=semithick,color=green]({291.0943 + 61.4057*x},{416.9198 + 82.5802*x});  
\addplot [>={Latex[length=1.5mm,width=0.5mm,angle'=25,open,round]},,domain=0:1,samples=2,style=semithick,color=green]({352.5 + 1.8149*x},{499.5 + 19.6377*x});  
\addplot [>={Latex[length=1.5mm,width=0.5mm,angle'=25,open,round]},,domain=19:24,samples=50,style=semithick,color=green]({49.309 + 7.5982*x + 0.44499*x^2},{4384.2449 + -343.13*x + 7.3528*x^2});  
\addplot [>={Latex[length=1.5mm,width=0.5mm,angle'=25,open,round]},,domain=0:1,samples=2,style=semithick,color=green]({487.9782 + 7.0218*x},{384.3409 + -3.3409*x});  
\addplot [>={Latex[length=1.5mm,width=0.5mm,angle'=25,open,round]},,domain=0:1,samples=2,style=semithick,color=green]({495 + 24.3192*x},{381 + 22.2647*x});  
\addplot [>={Latex[length=1.5mm,width=0.5mm,angle'=25,open,round]},,domain=25:27,samples=20,style=semithick,color=green]({-305.7726 + 37.5297*x + -0.18104*x^2},{4611.0245 + -370.2035*x + 8.0757*x^2});  
\addplot [>={Latex[length=1.5mm,width=0.5mm,angle'=25,open,round]},,domain=0:1,samples=2,style=semithick,color=green]({575.5504 + 13.9496*x},{502.733 + 11.767*x});  
\addplot [>={Latex[length=1.5mm,width=0.5mm,angle'=25,open,round]},,domain=0:1,samples=2,style=semithick,color=green]({589.5 + 12.1968*x},{514.5 + -19.0369*x});  
\addplot [>={Latex[length=1.5mm,width=0.5mm,angle'=25,open,round]},,domain=28:33,samples=50,style=semithick,color=green]({-170.1604 + 26.4309*x + 0.040552*x^2},{8741.8938 + -543.0847*x + 8.8775*x^2});  
\addplot [>={Latex[length=1.5mm,width=0.5mm,angle'=25,open,round]},,domain=0:1,samples=2,style=semithick,color=green]({746.2195 + 20.2805*x},{487.6697 + 31.3303*x});  
\addplot [>={Latex[length=1.5mm,width=0.5mm,angle'=25,open,round]},,domain=0:1,samples=2,style=semithick,color=green]({766.5 + 2.1517*x},{519 + -0.54275*x});  
\addplot [>={Latex[length=1.5mm,width=0.5mm,angle'=25,open,round]},,domain=34:39,samples=50,style=semithick,color=green]({-2774.734 + 174.2188*x + -2.0589*x^2},{10393.6429 + -544.1514*x + 7.4619*x^2});  
\addplot [>={Latex[length=1.5mm,width=0.5mm,angle'=25,open,round]},,domain=0:1,samples=2,style=semithick,color=green]({888.2582 + -3.6997*x},{521.2958 + 10.1052*x});  
\addplot [>={Latex[length=1.5mm,width=0.5mm,angle'=25,open,round]},,domain=39:41,samples=20,style=semithick,color=green]({1180.0958 + -0.74001*x + -0.17533*x^2},{12374.5178 + -583.6781*x + 7.1797*x^2});  
\addplot [>={Latex[length=1.5mm,width=0.5mm,angle'=25,open,round]},,domain=0:1,samples=2,style=semithick,color=green]({855.0257 + -18.0257*x},{512.7973 + 13.7027*x});  
\addplot [>={Latex[length=1.5mm,width=0.5mm,angle'=25,open,round]},,domain=0:1,samples=2,style=semithick,color=green]({837 + 2.2557*x},{526.5 + 3.7071*x});  
\addplot [>={Latex[length=1.5mm,width=0.5mm,angle'=25,open,round]},,domain=42:43,samples=10,style=semithick,color=green]({1419.5404 + -13.8163*x},{1083.2659 + -13.1681*x});  
\addplot [>={Latex[length=1.5mm,width=0.5mm,angle'=25,open,round]},,domain=0:1,samples=2,style=semithick,color=green]({825.4394 + -3.4394*x},{517.0391 + -1.0391*x});  
\addplot [>={Latex[length=1.5mm,width=0.5mm,angle'=25,open,round]},,domain=0:1,samples=2,style=semithick,color=green]({822 + -16.8792*x},{516 + 12.5461*x});  
\addplot [->,>={Latex[length=1.5mm,width=0.5mm,angle'=25,open,round]},,domain=44:46,samples=20,style=semithick,color=green]({-1391.2288 + 111.0891*x + -1.3903*x^2},{12237.4677 + -521.1568*x + 5.7965*x^2});  
\end{axis} 
\end{tikzpicture} 

  \noindent}} \\

\rotatebox{90}{\hspace{20pt} TbD~\cite{tbd}} &
\resizebox {!}{\sz\textwidth} {\resizebox {!}{0.2\textwidth} {\begin{tikzpicture} 
\begin{axis}[y dir=reverse, 
 xmin=1,xmax=960, 
 ymin=1,ymax=600, 
 xticklabels = \empty, yticklabels = \empty, 
 grid=none, axis equal image] 
\addplot graphics[xmin=1,xmax=960,ymin=1,ymax=600] {imgs/thumbnails/tbd/hit_tennis.jpg}; 
\addplot [<-,>={Latex[length=1.5mm,width=0.5mm,angle'=25,open,round]},,domain=0:1,samples=10,style=semithick,color={rgb,255:red,198; green,255; blue,0}]({645.013 + 88.5445*x + -16.4465*x^2},{526.1513 + 3.1922*x + -4.898*x^2});  
\addplot [<-,>={Latex[length=1.5mm,width=0.5mm,angle'=25,open,round]},,domain=0:1,samples=10,style=semithick,color={rgb,255:red,117; green,255; blue,0}]({588.5572 + 63.1017*x + -6.4654*x^2},{517.8041 + 30.2372*x + -29.8116*x^2});  
\addplot [<-,>={Latex[length=1.5mm,width=0.5mm,angle'=25,open,round]},,domain=0:1,samples=10,style=semithick,color={rgb,255:red,81; green,255; blue,0}]({523.2574 + 55.2528*x + 9.6477*x^2},{521.2588 + -6.1849*x + 6.7713*x^2});  
\addplot [<-,>={Latex[length=1.5mm,width=0.5mm,angle'=25,open,round]},,domain=0:1,samples=10,style=semithick,color={rgb,255:red,83; green,255; blue,0}]({463.2762 + 54.2548*x + 9.0905*x^2},{523.8773 + -11.92*x + 8.1485*x^2});  
\addplot [<-,>={Latex[length=1.5mm,width=0.5mm,angle'=25,open,round]},,domain=0:1,samples=10,style=semithick,color={rgb,255:red,110; green,255; blue,0}]({408.5974 + 55.9656*x + -0.041389*x^2},{516.2679 + 27.6336*x + -27.3791*x^2});  
\addplot [<-,>={Latex[length=1.5mm,width=0.5mm,angle'=25,open,round]},,domain=0:1,samples=10,style=semithick,color={rgb,255:red,92; green,255; blue,0}]({348.1348 + 64.3373*x + -4.5262*x^2},{515.0168 + 8.8776*x + -7.0126*x^2});  
\addplot [<-,>={Latex[length=1.5mm,width=0.5mm,angle'=25,open,round]},,domain=0:1,samples=2,style=semithick,color={rgb,255:red,83; green,255; blue,0}]({292.6283 + 58.5023*x},{512.8202 + 4.4698*x});  
\addplot [<-,>={Latex[length=1.5mm,width=0.5mm,angle'=25,open,round]},,domain=0:1,samples=10,style=semithick,color={rgb,255:red,87; green,255; blue,0}]({235.3512 + 56.4627*x + 4.0677*x^2},{513.6489 + -3.8263*x + 6.3102*x^2});  
\addplot [->,>={Latex[length=1.5mm,width=0.5mm,angle'=25,open,round]},,domain=0:1,samples=10,style=semithick,color={rgb,255:red,140; green,255; blue,0}]({240.1649 + -33.9807*x + -23.2887*x^2},{515.4697 + -13.8948*x + 12.2179*x^2});  
\addplot [<-,>={Latex[length=1.5mm,width=0.5mm,angle'=25,open,round]},,domain=0:1,samples=2,style=semithick,color={rgb,255:red,137; green,255; blue,0}]({125.971 + 60.0369*x},{513.1667 + -1.3063*x});  
\addplot [<-,>={Latex[length=1.5mm,width=0.5mm,angle'=25,open,round]},,domain=0:1,samples=10,style=semithick,color={rgb,255:red,136; green,255; blue,0}]({73.4934 + 62.7067*x + -4.2237*x^2},{511.3964 + -4.2248*x + 1.6156*x^2});  
\addplot [->,>={Latex[length=1.5mm,width=0.5mm,angle'=25,open,round]},,domain=0:1,samples=10,style=semithick,color={rgb,255:red,77; green,255; blue,0}]({72.535 + 48.8196*x + -5.4008*x^2},{512.0376 + -50.8722*x + 34.2244*x^2});  
\addplot [->,>={Latex[length=1.425mm,width=0.5mm,angle'=25,open,round]},,domain=0:1,samples=10,style=semithick,color={rgb,255:red,59; green,255; blue,0}]({115.3701 + 43.0179*x + -0.88221*x^2},{492.3806 + -6.3464*x + 10.3761*x^2});  
\addplot [->,>={Latex[length=0.66816mm,width=0.5mm,angle'=25,open,round]},,domain=0:1,samples=2,style=semithick,color={rgb,255:red,137; green,255; blue,0}]({171.3906 + 13.2481*x},{511.6651 + -15.0428*x});  
\addplot [<-,>={Latex[length=0.51334mm,width=0.5mm,angle'=25,open,round]},,domain=0:1,samples=10,style=semithick,color={rgb,255:red,51; green,255; blue,0}]({199.2515 + -15.5151*x + 0.61283*x^2},{493.95 + 0.83415*x + 2.7061*x^2});  
\addplot [->,>={Latex[length=0.64027mm,width=0.5mm,angle'=25,open,round]},,domain=0:1,samples=2,style=semithick,color={rgb,255:red,60; green,255; blue,0}]({200.0009 + 15.1296*x},{494.3597 + 11.8343*x});  
\addplot [->,>={Latex[length=0.91662mm,width=0.5mm,angle'=25,open,round]},,domain=0:1,samples=10,style=semithick,color={rgb,255:red,94; green,255; blue,0}]({211.8407 + 16.4215*x + 6.6793*x^2},{506.3572 + -29.301*x + 21.3741*x^2});  
\addplot [->,>={Latex[length=0.85244mm,width=0.5mm,angle'=25,open,round]},,domain=0:1,samples=2,style=semithick,color={rgb,255:red,95; green,255; blue,0}]({229.8755 + 23.2824*x},{506.3263 + -10.5792*x});  
\addplot [->,>={Latex[length=0.72797mm,width=0.5mm,angle'=25,open,round]},,domain=0:1,samples=2,style=semithick,color={rgb,255:red,82; green,255; blue,0}]({253.5444 + 21.0613*x},{496.3383 + 5.7764*x});  
\addplot [->,>={Latex[length=0.83784mm,width=0.5mm,angle'=25,open,round]},,domain=0:1,samples=10,style=semithick,color={rgb,255:red,91; green,255; blue,0}]({267.9989 + 23.8144*x + 0.82243*x^2},{500.6286 + -9.2483*x + 8.5696*x^2});  
\addplot [->,>={Latex[length=0.68167mm,width=0.5mm,angle'=25,open,round]},,domain=0:1,samples=10,style=semithick,color={rgb,255:red,76; green,255; blue,0}]({287.5764 + 18.6837*x + 0.77069*x^2},{501.5114 + -8.9208*x + 2.8504*x^2});  
\addplot [->,>={Latex[length=0.79808mm,width=0.5mm,angle'=25,open,round]},,domain=0:1,samples=10,style=semithick,color={rgb,255:red,88; green,255; blue,0}]({303.0697 + 22.1148*x + 0.65873*x^2},{500.7208 + -14.7313*x + 12.0305*x^2});  
\addplot [->,>={Latex[length=0.6804mm,width=0.5mm,angle'=25,open,round]},,domain=0:1,samples=10,style=semithick,color={rgb,255:red,90; green,255; blue,0}]({319.4462 + 18.8757*x + 0.65989*x^2},{499.2751 + -8.1908*x + 2.4585*x^2});  
\addplot [->,>={Latex[length=0.63036mm,width=0.5mm,angle'=25,open,round]},,domain=0:1,samples=2,style=semithick,color={rgb,255:red,79; green,255; blue,0}]({337.1378 + 18.3671*x},{496.5224 + -4.5026*x});  
\addplot [->,>={Latex[length=0.61296mm,width=0.5mm,angle'=25,open,round]},,domain=0:1,samples=2,style=semithick,color={rgb,255:red,68; green,255; blue,0}]({353.7122 + 18.2747*x},{493.9291 + -2.0461*x});  
\addplot [->,>={Latex[length=0.6057mm,width=0.5mm,angle'=25,open,round]},,domain=0:1,samples=2,style=semithick,color={rgb,255:red,76; green,255; blue,0}]({370.3257 + 18.1225*x},{492.6206 + -1.3277*x});  
\end{axis} 
\end{tikzpicture} 

  \noindent}} & 
\resizebox {!}{\sz\textwidth} {\resizebox {!}{0.2\textwidth} {\begin{tikzpicture} 
\begin{axis}[y dir=reverse, 
 xmin=1,xmax=960, 
 ymin=1,ymax=600, 
 xticklabels = \empty, yticklabels = \empty, 
 grid=none, axis equal image] 
\addplot graphics[xmin=1,xmax=960,ymin=1,ymax=600] {imgs/thumbnails/tbd/roll_golf.jpg}; 
\addplot [->,>={Latex[length=1.5mm,width=0.5mm,angle'=25,open,round]},,domain=0:1,samples=2,style=semithick,color={rgb,255:red,248; green,255; blue,0}]({252.3463 + 22.145*x},{564.5278 + -77.4591*x});  
\addplot [->,>={Latex[length=1.5mm,width=0.5mm,angle'=25,open,round]},,domain=0:1,samples=2,style=semithick,color={rgb,255:red,52; green,255; blue,0}]({281.1912 + 20.8012*x},{454.6955 + -73.5562*x});  
\addplot [->,>={Latex[length=1.5mm,width=0.5mm,angle'=25,open,round]},,domain=0:1,samples=2,style=semithick,color={rgb,255:red,45; green,255; blue,0}]({302.446 + 22.8349*x},{379.7656 + -81.3271*x});  
\addplot [->,>={Latex[length=1.5mm,width=0.5mm,angle'=25,open,round]},,domain=0:1,samples=2,style=semithick,color={rgb,255:red,41; green,255; blue,0}]({325.4789 + 20.9672*x},{295.9031 + -76.9178*x});  
\addplot [->,>={Latex[length=1.5mm,width=0.5mm,angle'=25,open,round]},,domain=0:1,samples=2,style=semithick,color={rgb,255:red,53; green,255; blue,0}]({346.9097 + 17.8053*x},{216.4672 + -72.7816*x});  
\addplot [->,>={Latex[length=1.5mm,width=0.5mm,angle'=25,open,round]},,domain=0:1,samples=2,style=semithick,color={rgb,255:red,65; green,255; blue,0}]({364.6613 + 16.7447*x},{141.3993 + -69.6623*x});  
\addplot [<-,>={Latex[length=0.98231mm,width=0.5mm,angle'=25,open,round]},,domain=0:1,samples=2,style=semithick,color={rgb,255:red,154; green,255; blue,0}]({391.946 + -7.7355*x},{92.8786 + -28.4359*x});  
\addplot [<-,>={Latex[length=0.7604mm,width=0.5mm,angle'=25,open,round]},,domain=0:1,samples=2,style=semithick,color={rgb,255:red,32; green,255; blue,0}]({399.5865 + -7.8066*x},{115.8327 + -21.4346*x});  
\addplot [<-,>={Latex[length=0.66924mm,width=0.5mm,angle'=25,open,round]},,domain=0:1,samples=2,style=semithick,color={rgb,255:red,25; green,255; blue,0}]({408.9029 + -7.7236*x},{139.7489 + -18.532*x});  
\addplot [<-,>={Latex[length=0.63136mm,width=0.5mm,angle'=25,open,round]},,domain=0:1,samples=2,style=semithick,color={rgb,255:red,30; green,255; blue,0}]({413.7741 + -4.9219*x},{160.5608 + -18.2901*x});  
\addplot [<-,>={Latex[length=0.61837mm,width=0.5mm,angle'=25,open,round]},,domain=0:1,samples=2,style=semithick,color={rgb,255:red,14; green,255; blue,0}]({417.3852 + -4.187*x},{181.4107 + -18.0725*x});  
\addplot [<-,>={Latex[length=0.60793mm,width=0.5mm,angle'=25,open,round]},,domain=0:1,samples=2,style=semithick,color={rgb,255:red,34; green,255; blue,0}]({420.1112 + -3.5993*x},{200.9776 + -17.8794*x});  
\addplot [<-,>={Latex[length=0.55878mm,width=0.5mm,angle'=25,open,round]},,domain=0:1,samples=2,style=semithick,color={rgb,255:red,10; green,255; blue,0}]({422.1548 + -2.514*x},{220.6282 + -16.5737*x});  
\addplot [<-,>={Latex[length=0.52047mm,width=0.5mm,angle'=25,open,round]},,domain=0:1,samples=2,style=semithick,color={rgb,255:red,30; green,255; blue,0}]({425.4829 + -3.3253*x},{238.6127 + -15.256*x});  
\addplot [<-,>={Latex[length=0.52177mm,width=0.5mm,angle'=25,open,round]},,domain=0:1,samples=2,style=semithick,color={rgb,255:red,22; green,255; blue,0}]({430.0465 + -4.9539*x},{253.6513 + -14.8485*x});  
\end{axis} 
\end{tikzpicture} 

  \noindent}} & 
\resizebox {!}{\sz\textwidth} {\resizebox {!}{0.2\textwidth} {\begin{tikzpicture} 
\begin{axis}[y dir=reverse, 
 xmin=1,xmax=1920, 
 ymin=1,ymax=1080, 
 xticklabels = \empty, yticklabels = \empty, 
 grid=none, axis equal image] 
\addplot graphics[xmin=1,xmax=1920,ymin=1,ymax=1080] {imgs/thumbnails/tbd/tennis.jpg}; 
\addplot [<-,>={Latex[length=1.1401mm,width=0.5mm,angle'=25,open,round]},,domain=0:1,samples=2,style=semithick,color={rgb,255:red,50; green,255; blue,0}]({1554.3065 + 17.8483*x},{354.6145 + 14.1907*x});  
\addplot [<-,>={Latex[length=1.4008mm,width=0.5mm,angle'=25,open,round]},,domain=0:1,samples=2,style=semithick,color={rgb,255:red,52; green,255; blue,0}]({1530.0803 + 23.1266*x},{337.8826 + 15.8148*x});  
\addplot [<-,>={Latex[length=1.3892mm,width=0.5mm,angle'=25,open,round]},,domain=0:1,samples=2,style=semithick,color={rgb,255:red,62; green,255; blue,0}]({1505.1614 + 24.103*x},{323.7184 + 13.8204*x});  
\addplot [<-,>={Latex[length=1.3865mm,width=0.5mm,angle'=25,open,round]},,domain=0:1,samples=2,style=semithick,color={rgb,255:red,58; green,255; blue,0}]({1478.8547 + 25.0974*x},{311.3092 + 11.7927*x});  
\addplot [<-,>={Latex[length=1.4396mm,width=0.5mm,angle'=25,open,round]},,domain=0:1,samples=2,style=semithick,color={rgb,255:red,73; green,255; blue,0}]({1451.0475 + 26.9079*x},{300.8751 + 10.242*x});  
\addplot [<-,>={Latex[length=1.4701mm,width=0.5mm,angle'=25,open,round]},,domain=0:1,samples=2,style=semithick,color={rgb,255:red,72; green,255; blue,0}]({1421.8055 + 28.2052*x},{292.6608 + 8.3029*x});  
\addplot [<-,>={Latex[length=1.4906mm,width=0.5mm,angle'=25,open,round]},,domain=0:1,samples=2,style=semithick,color={rgb,255:red,80; green,255; blue,0}]({1392.0531 + 29.2194*x},{286.7376 + 5.9162*x});  
\addplot [<-,>={Latex[length=1.5mm,width=0.5mm,angle'=25,open,round]},,domain=0:1,samples=2,style=semithick,color={rgb,255:red,67; green,255; blue,0}]({1360.7914 + 30.2887*x},{282.8037 + 3.5036*x});  
\addplot [<-,>={Latex[length=1.5mm,width=0.5mm,angle'=25,open,round]},,domain=0:1,samples=2,style=semithick,color={rgb,255:red,55; green,255; blue,0}]({1328.0022 + 32.1236*x},{280.9671 + 2.1655*x});  
\addplot [<-,>={Latex[length=1.5mm,width=0.5mm,angle'=25,open,round]},,domain=0:1,samples=10,style=semithick,color={rgb,255:red,86; green,255; blue,0}]({1293.7293 + 27.8019*x + 6.4472*x^2},{282.4861 + -5.2349*x + 4.0921*x^2});  
\addplot [<-,>={Latex[length=1.5mm,width=0.5mm,angle'=25,open,round]},,domain=0:1,samples=2,style=semithick,color={rgb,255:red,96; green,255; blue,0}]({1256.9181 + 38.0495*x},{287.321 + -4.5896*x});  
\addplot [<-,>={Latex[length=1.5mm,width=0.5mm,angle'=25,open,round]},,domain=0:1,samples=2,style=semithick,color={rgb,255:red,70; green,255; blue,0}]({1222.9258 + 36.8447*x},{293.5999 + -6.8371*x});  
\addplot [<-,>={Latex[length=1.5mm,width=0.5mm,angle'=25,open,round]},,domain=0:1,samples=10,style=semithick,color={rgb,255:red,55; green,255; blue,0}]({1187.753 + 35.3092*x + 1.0623*x^2},{303.1967 + -10.8588*x + 1.2186*x^2});  
\addplot [<-,>={Latex[length=1.5mm,width=0.5mm,angle'=25,open,round]},,domain=0:1,samples=2,style=semithick,color={rgb,255:red,71; green,255; blue,0}]({1148.8232 + 39.0161*x},{315.4674 + -12.9515*x});  
\addplot [<-,>={Latex[length=1.5mm,width=0.5mm,angle'=25,open,round]},,domain=0:1,samples=2,style=semithick,color={rgb,255:red,62; green,255; blue,0}]({1111.7637 + 38.108*x},{330.4279 + -15.7384*x});  
\addplot [->,>={Latex[length=1.5mm,width=0.5mm,angle'=25,open,round]},,domain=0:1,samples=10,style=semithick,color={rgb,255:red,82; green,255; blue,0}]({1111.2759 + -43.6949*x + 5.4095*x^2},{330.6264 + 19.4294*x + -1.0679*x^2});  
\addplot [<-,>={Latex[length=1.5mm,width=0.5mm,angle'=25,open,round]},,domain=0:1,samples=10,style=semithick,color={rgb,255:red,97; green,255; blue,0}]({1032.7201 + 33.4228*x + 7.1702*x^2},{371.1652 + -20.7935*x + -1.7532*x^2});  
\addplot [<-,>={Latex[length=1.5mm,width=0.5mm,angle'=25,open,round]},,domain=0:1,samples=2,style=semithick,color={rgb,255:red,136; green,255; blue,0}]({1001.4003 + 23.7524*x},{394.4872 + -19.5182*x});  
\addplot [<-,>={Latex[length=1.5mm,width=0.5mm,angle'=25,open,round]},,domain=0:1,samples=2,style=semithick,color={rgb,255:red,34; green,255; blue,0}]({955.9312 + 37.7614*x},{421.9014 + -26.342*x});  
\addplot [->,>={Latex[length=1.5mm,width=0.5mm,angle'=25,open,round]},,domain=0:1,samples=10,style=semithick,color={rgb,255:red,255; green,226; blue,0}]({945.0427 + 25.8306*x + -53.4991*x^2},{411.8279 + 29.1239*x + 7.8928*x^2});  
\addplot [<-,>={Latex[length=1.5mm,width=0.5mm,angle'=25,open,round]},,domain=0:1,samples=10,style=semithick,color={rgb,255:red,255; green,3; blue,0}]({830.2205 + 51.1679*x + 13.3339*x^2},{485.6358 + -17.0557*x + -8.2383*x^2});  
\addplot [<-,>={Latex[length=1.5mm,width=0.5mm,angle'=25,open,round]},,domain=0:1,samples=2,style=semithick,color=red]({786.1773 + 32.9919*x},{495.1576 + -5.0524*x});  
\addplot [<-,>={Latex[length=1.5mm,width=0.5mm,angle'=25,open,round]},,domain=0:1,samples=2,style=semithick,color={rgb,255:red,66; green,255; blue,0}]({812.9434 + 32.4653*x},{546.9435 + -32.5336*x});  
\addplot [<-,>={Latex[length=1.5mm,width=0.5mm,angle'=25,open,round]},,domain=0:1,samples=2,style=semithick,color={rgb,255:red,46; green,255; blue,0}]({780.3077 + 31.2799*x},{582.2858 + -33.669*x});  
\addplot [<-,>={Latex[length=1.5mm,width=0.5mm,angle'=25,open,round]},,domain=0:1,samples=2,style=semithick,color={rgb,255:red,54; green,255; blue,0}]({749.4065 + 29.5903*x},{618.4909 + -34.4936*x});  
\addplot [->,>={Latex[length=1.2877mm,width=0.5mm,angle'=25,open,round]},,domain=0:1,samples=2,style=semithick,color={rgb,255:red,157; green,255; blue,0}]({742.9501 + -13.5879*x},{628.6521 + -21.8771*x});  
\addplot [->,>={Latex[length=1.4841mm,width=0.5mm,angle'=25,open,round]},,domain=0:1,samples=2,style=semithick,color={rgb,255:red,100; green,255; blue,0}]({729.6199 + -16.0406*x},{605.6018 + -24.9739*x});  
\addplot [->,>={Latex[length=1.3598mm,width=0.5mm,angle'=25,open,round]},,domain=0:1,samples=2,style=semithick,color={rgb,255:red,73; green,255; blue,0}]({712.9099 + -15.48*x},{579.0623 + -22.36*x});  
\addplot [->,>={Latex[length=1.3102mm,width=0.5mm,angle'=25,open,round]},,domain=0:1,samples=2,style=semithick,color={rgb,255:red,87; green,255; blue,0}]({696.8278 + -15.7877*x},{555.13 + -20.9151*x});  
\addplot [->,>={Latex[length=1.1988mm,width=0.5mm,angle'=25,open,round]},,domain=0:1,samples=2,style=semithick,color={rgb,255:red,79; green,255; blue,0}]({681.0173 + -15.6626*x},{532.8479 + -18.1538*x});  
\addplot [<-,>={Latex[length=1.1672mm,width=0.5mm,angle'=25,open,round]},,domain=0:1,samples=2,style=semithick,color={rgb,255:red,93; green,255; blue,0}]({649.1505 + 16.2023*x},{496.8549 + 16.8049*x});  
\addplot [<-,>={Latex[length=1.0664mm,width=0.5mm,angle'=25,open,round]},,domain=0:1,samples=2,style=semithick,color={rgb,255:red,61; green,255; blue,0}]({633.9286 + 15.8164*x},{481.9736 + 14.3083*x});  
\addplot [<-,>={Latex[length=0.93864mm,width=0.5mm,angle'=25,open,round]},,domain=0:1,samples=2,style=semithick,color={rgb,255:red,73; green,255; blue,0}]({618.9831 + 14.8404*x},{469.731 + 11.4968*x});  
\addplot [<-,>={Latex[length=1.0392mm,width=0.5mm,angle'=25,open,round]},,domain=0:1,samples=2,style=semithick,color={rgb,255:red,179; green,255; blue,0}]({599.3631 + 16.3898*x},{454.5345 + 12.7823*x});  
\addplot [<-,>={Latex[length=1.0792mm,width=0.5mm,angle'=25,open,round]},,domain=0:1,samples=2,style=semithick,color={rgb,255:red,255; green,132; blue,0}]({578.4765 + 17.1028*x},{438.381 + 13.1654*x});  
\addplot [<-,>={Latex[length=1.1491mm,width=0.5mm,angle'=25,open,round]},,domain=0:1,samples=2,style=semithick,color={rgb,255:red,255; green,5; blue,0}]({556.4617 + 18.1821*x},{421.4029 + 14.0577*x});  
\end{axis} 
\end{tikzpicture} 

  \noindent}} & 
\resizebox {!}{\sz\textwidth} {\resizebox {!}{0.2\textwidth} {\begin{tikzpicture} 
\begin{axis}[y dir=reverse, 
 xmin=1,xmax=960, 
 ymin=1,ymax=600, 
 xticklabels = \empty, yticklabels = \empty, 
 grid=none, axis equal image] 
\addplot graphics[xmin=1,xmax=960,ymin=1,ymax=600] {imgs/thumbnails/tbd/throw_tennis.jpg}; 
\addplot [<-,>={Latex[length=1.5mm,width=0.5mm,angle'=25,open,round]},,domain=0:1,samples=2,style=semithick,color={rgb,255:red,113; green,255; blue,0}]({769.502 + 77.4291*x},{297.002 + -77.0735*x});  
\addplot [<-,>={Latex[length=1.5mm,width=0.5mm,angle'=25,open,round]},,domain=0:1,samples=2,style=semithick,color={rgb,255:red,56; green,255; blue,0}]({679.7764 + 84.6658*x},{400.9474 + -98.4272*x});  
\addplot [<-,>={Latex[length=1.5mm,width=0.5mm,angle'=25,open,round]},,domain=0:1,samples=10,style=semithick,color={rgb,255:red,73; green,255; blue,0}]({595.8696 + 91.8952*x + -11.8955*x^2},{515.8494 + -127.5539*x + 19.228*x^2});  
\addplot [->,>={Latex[length=1.5mm,width=0.5mm,angle'=25,open,round]},,domain=0:1,samples=2,style=semithick,color={rgb,255:red,84; green,255; blue,0}]({590.2227 + -70.4596*x},{515.1313 + -86.7511*x});  
\addplot [<-,>={Latex[length=1.5mm,width=0.5mm,angle'=25,open,round]},,domain=0:1,samples=10,style=semithick,color={rgb,255:red,43; green,255; blue,0}]({441.2487 + 74.0552*x + -1.0922*x^2},{352.2135 + 62.3832*x + 8.6702*x^2});  
\addplot [<-,>={Latex[length=1.5mm,width=0.5mm,angle'=25,open,round]},,domain=0:1,samples=2,style=semithick,color={rgb,255:red,61; green,255; blue,0}]({367.983 + 72.6916*x},{294.8567 + 54.5788*x});  
\addplot [<-,>={Latex[length=1.5mm,width=0.5mm,angle'=25,open,round]},,domain=0:1,samples=2,style=semithick,color={rgb,255:red,77; green,255; blue,0}]({291.4055 + 75.571*x},{254.2222 + 39.4049*x});  
\addplot [<-,>={Latex[length=1.5mm,width=0.5mm,angle'=25,open,round]},,domain=0:1,samples=2,style=semithick,color={rgb,255:red,42; green,255; blue,0}]({214.4452 + 74.0089*x},{232.7132 + 19.0199*x});  
\addplot [<-,>={Latex[length=1.5mm,width=0.5mm,angle'=25,open,round]},,domain=0:1,samples=2,style=semithick,color={rgb,255:red,80; green,255; blue,0}]({139.6245 + 77.9128*x},{227.3478 + 4.5074*x});  
\addplot [<-,>={Latex[length=1.5mm,width=0.5mm,angle'=25,open,round]},,domain=0:1,samples=10,style=semithick,color={rgb,255:red,50; green,255; blue,0}]({69.2085 + 68.6378*x + 1.6536*x^2},{242.1527 + -22.8971*x + 10.3714*x^2});  
\addplot [->,>={Latex[length=1.3665mm,width=0.5mm,angle'=25,open,round]},,domain=0:1,samples=2,style=semithick,color={rgb,255:red,129; green,255; blue,0}]({30.3938 + 36.3493*x},{261.8783 + -18.9531*x});  
\addplot [->,>={Latex[length=1.5mm,width=0.5mm,angle'=25,open,round]},,domain=0:1,samples=2,style=semithick,color={rgb,255:red,23; green,255; blue,0}]({64.305 + 51.6766*x},{255.6383 + 15.7846*x});  
\addplot [->,>={Latex[length=1.5mm,width=0.5mm,angle'=25,open,round]},,domain=0:1,samples=10,style=semithick,color={rgb,255:red,54; green,255; blue,0}]({116.7032 + 61.5336*x + -4.3684*x^2},{272.2699 + 23.1777*x + 10.706*x^2});  
\addplot [->,>={Latex[length=1.5mm,width=0.5mm,angle'=25,open,round]},,domain=0:1,samples=2,style=semithick,color={rgb,255:red,50; green,255; blue,0}]({175.109 + 55.7538*x},{304.957 + 47.7033*x});  
\addplot [<-,>={Latex[length=1.5mm,width=0.5mm,angle'=25,open,round]},,domain=0:1,samples=2,style=semithick,color={rgb,255:red,47; green,255; blue,0}]({288.4268 + -55.9352*x},{421.1362 + -65.6291*x});  
\addplot [->,>={Latex[length=1.5mm,width=0.5mm,angle'=25,open,round]},,domain=0:1,samples=10,style=semithick,color={rgb,255:red,116; green,255; blue,0}]({289.6456 + 43.0125*x + 15.485*x^2},{425.8723 + 51.2815*x + 33.6897*x^2});  
\addplot [->,>={Latex[length=1.5mm,width=0.5mm,angle'=25,open,round]},,domain=0:1,samples=2,style=semithick,color={rgb,255:red,56; green,255; blue,0}]({354.745 + 23.3464*x},{518.5542 + -49.7157*x});  
\addplot [->,>={Latex[length=1.5mm,width=0.5mm,angle'=25,open,round]},,domain=0:1,samples=2,style=semithick,color={rgb,255:red,32; green,255; blue,0}]({377.3389 + 27.0497*x},{465.5113 + -44.3839*x});  
\addplot [->,>={Latex[length=1.2736mm,width=0.5mm,angle'=25,open,round]},,domain=0:1,samples=2,style=semithick,color={rgb,255:red,24; green,255; blue,0}]({404.4616 + 27.3837*x},{417.9467 + -26.6471*x});  
\addplot [->,>={Latex[length=1.0497mm,width=0.5mm,angle'=25,open,round]},,domain=0:1,samples=2,style=semithick,color={rgb,255:red,32; green,255; blue,0}]({431.4289 + 29.1387*x},{390.1063 + -11.9414*x});  
\addplot [->,>={Latex[length=0.99413mm,width=0.5mm,angle'=25,open,round]},,domain=0:1,samples=10,style=semithick,color={rgb,255:red,25; green,255; blue,0}]({461.3835 + 29.1146*x + -1.4769*x^2},{380.4724 + -10.4826*x + 16.5526*x^2});  
\addplot [->,>={Latex[length=1.1711mm,width=0.5mm,angle'=25,open,round]},,domain=0:1,samples=2,style=semithick,color={rgb,255:red,28; green,255; blue,0}]({491.0179 + 27.5985*x},{383.3426 + 21.7402*x});  
\addplot [->,>={Latex[length=1.5mm,width=0.5mm,angle'=25,open,round]},,domain=0:1,samples=10,style=semithick,color={rgb,255:red,36; green,255; blue,0}]({517.0835 + 31.8111*x + -3.4831*x^2},{406.938 + 31.9636*x + 6.9215*x^2});  
\addplot [<-,>={Latex[length=1.5mm,width=0.5mm,angle'=25,open,round]},,domain=0:1,samples=2,style=semithick,color={rgb,255:red,43; green,255; blue,0}]({574.4917 + -27.7115*x},{504.7703 + -56.6519*x});  
\addplot [->,>={Latex[length=1.2064mm,width=0.5mm,angle'=25,open,round]},,domain=0:1,samples=2,style=semithick,color={rgb,255:red,118; green,255; blue,0}]({581.7972 + 20.0682*x},{528.8643 + -30.1202*x});  
\addplot [->,>={Latex[length=1.4603mm,width=0.5mm,angle'=25,open,round]},,domain=0:1,samples=2,style=semithick,color={rgb,255:red,40; green,255; blue,0}]({602.2376 + 26.7274*x},{495.9129 + -34.7099*x});  
\addplot [->,>={Latex[length=1.0945mm,width=0.5mm,angle'=25,open,round]},,domain=0:1,samples=10,style=semithick,color={rgb,255:red,24; green,255; blue,0}]({631.6198 + 21.3767*x + 5.4443*x^2},{459.0904 + -25.3496*x + 7.1243*x^2});  
\addplot [<-,>={Latex[length=0.90653mm,width=0.5mm,angle'=25,open,round]},,domain=0:1,samples=10,style=semithick,color={rgb,255:red,38; green,255; blue,0}]({687.3445 + -27.5626*x + 1.6255*x^2},{439.8926 + -11.2184*x + 13.5934*x^2});  
\addplot [->,>={Latex[length=1.0408mm,width=0.5mm,angle'=25,open,round]},,domain=0:1,samples=10,style=semithick,color={rgb,255:red,30; green,255; blue,0}]({688.4758 + 30.1372*x + -2.8747*x^2},{439.7186 + 6.5343*x + 7.9114*x^2});  
\addplot [<-,>={Latex[length=1.4448mm,width=0.5mm,angle'=25,open,round]},,domain=0:1,samples=2,style=semithick,color={rgb,255:red,34; green,255; blue,0}]({743.1772 + -27.1326*x},{488.1604 + -33.8017*x});  
\addplot [->,>={Latex[length=1.5mm,width=0.5mm,angle'=25,open,round]},,domain=0:1,samples=10,style=semithick,color={rgb,255:red,44; green,255; blue,0}]({742.1305 + 27.3783*x + -2.646*x^2},{488.2132 + 39.216*x + 2.6365*x^2});  
\addplot [->,>={Latex[length=1.4708mm,width=0.5mm,angle'=25,open,round]},,domain=0:1,samples=10,style=semithick,color={rgb,255:red,38; green,255; blue,0}]({768.8735 + 28.3971*x + 0.49244*x^2},{523.0727 + -42.2063*x + 9.055*x^2});  
\addplot [->,>={Latex[length=1.0547mm,width=0.5mm,angle'=25,open,round]},,domain=0:1,samples=2,style=semithick,color={rgb,255:red,16; green,255; blue,0}]({797.7462 + 26.9557*x},{488.5872 + -16.572*x});  
\addplot [->,>={Latex[length=0.90009mm,width=0.5mm,angle'=25,open,round]},,domain=0:1,samples=2,style=semithick,color={rgb,255:red,38; green,255; blue,0}]({826.5003 + 27.0028*x},{472.3494 + 0.05228*x});  
\addplot [->,>={Latex[length=1.0404mm,width=0.5mm,angle'=25,open,round]},,domain=0:1,samples=2,style=semithick,color={rgb,255:red,42; green,255; blue,0}]({853.9395 + 25.5624*x},{474.8727 + 17.9109*x});  
\addplot [->,>={Latex[length=0.77977mm,width=0.5mm,angle'=25,open,round]},,domain=0:1,samples=2,style=semithick,color={rgb,255:red,106; green,255; blue,0}]({893.4649 + -13.9354*x},{510.7842 + -18.7894*x});  
\addplot [->,>={Latex[length=0.56997mm,width=0.5mm,angle'=25,open,round]},,domain=0:1,samples=2,style=semithick,color={rgb,255:red,63; green,255; blue,0}]({874.0218 + -4.1715*x},{529.2429 + -16.5826*x});  
\addplot [<-,>={Latex[length=0.5mm,width=0.5mm,angle'=25,open,round]},,domain=0:1,samples=2,style=semithick,color={rgb,255:red,31; green,255; blue,0}]({853.2797 + 12.7081*x},{509.4458 + 2.9601*x});  
\addplot [<-,>={Latex[length=0.70748mm,width=0.5mm,angle'=25,open,round]},,domain=0:1,samples=2,style=semithick,color={rgb,255:red,34; green,255; blue,0}]({835.3605 + 14.9261*x},{526.3783 + -15.0893*x});  
\addplot [<-,>={Latex[length=0.65098mm,width=0.5mm,angle'=25,open,round]},,domain=0:1,samples=10,style=semithick,color={rgb,255:red,31; green,255; blue,0}]({822.5105 + 17.0374*x + -1.4557*x^2},{516.6798 + 9.6798*x + 2.0065*x^2});  
\addplot [<-,>={Latex[length=0.63472mm,width=0.5mm,angle'=25,open,round]},,domain=0:1,samples=10,style=semithick,color={rgb,255:red,29; green,255; blue,0}]({805.1919 + 14.5981*x + 1.6977*x^2},{526.014 + -12.3616*x + 2.6822*x^2});  
\addplot [<-,>={Latex[length=0.55862mm,width=0.5mm,angle'=25,open,round]},,domain=0:1,samples=2,style=semithick,color={rgb,255:red,35; green,255; blue,0}]({790.5595 + 14.6975*x},{521.8913 + 8.0521*x});  
\addplot [<-,>={Latex[length=0.57989mm,width=0.5mm,angle'=25,open,round]},,domain=0:1,samples=2,style=semithick,color={rgb,255:red,28; green,255; blue,0}]({774.9455 + 15.3985*x},{529.7983 + -8.0951*x});  
\end{axis} 
\end{tikzpicture} 

  \noindent}} \\

\rotatebox{90}{\hspace{8pt} TbD-NC~\cite{tbdnc}} &
\resizebox {!}{\sz\textwidth} {\resizebox {!}{0.2\textwidth} {\begin{tikzpicture} 
\begin{axis}[y dir=reverse, 
 xmin=1,xmax=960, 
 ymin=1,ymax=600, 
 xticklabels = \empty, yticklabels = \empty, 
 grid=none, axis equal image] 
\addplot graphics[xmin=1,xmax=960,ymin=1,ymax=600] {imgs/thumbnails/tbd/hit_tennis.jpg}; 
\addplot [>={Latex[length=1.5mm,width=0.5mm,angle'=25,open,round]},,domain=1:5,samples=40,style=semithick,color=green]({1077.6011 + -72.098*x},{515.9165 + 1.7058*x});  
\addplot [>={Latex[length=1.5mm,width=0.5mm,angle'=25,open,round]},,domain=5:6,samples=10,style=semithick,color=green]({1077.6011 + -72.098*x},{515.9165 + 1.7058*x});  
\addplot [>={Latex[length=1.5mm,width=0.5mm,angle'=25,open,round]},,domain=0:1,samples=2,style=semithick,color=green]({645.013 + -33.013*x},{526.1513 + -1.1513*x});  
\addplot [>={Latex[length=1.5mm,width=0.5mm,angle'=25,open,round]},,domain=0:1,samples=2,style=semithick,color=green]({612 + -23.8422*x},{525 + -3.1547*x});  
\addplot [>={Latex[length=1.5mm,width=0.5mm,angle'=25,open,round]},,domain=7:9,samples=20,style=semithick,color=green]({1074.2305 + -74.882*x + 0.77757*x^2},{652.0085 + -33.8475*x + 2.179*x^2});  
\addplot [>={Latex[length=1.5mm,width=0.5mm,angle'=25,open,round]},,domain=0:1,samples=2,style=semithick,color=green]({463.2762 + -31.2762*x},{523.8773 + -0.37727*x});  
\addplot [>={Latex[length=1.5mm,width=0.5mm,angle'=25,open,round]},,domain=0:1,samples=2,style=semithick,color=green]({432 + -24.054*x},{523.5 + -6.6182*x});  
\addplot [>={Latex[length=1.5mm,width=0.5mm,angle'=25,open,round]},,domain=10:16,samples=60,style=semithick,color=green]({1004.0891 + -62.0345*x + 0.24201*x^2},{532.0524 + -1.8938*x + 0.037677*x^2});  
\addplot [>={Latex[length=1.5mm,width=0.5mm,angle'=25,open,round]},,domain=0:1,samples=2,style=semithick,color=green]({73.4934 + -0.95844*x},{511.3964 + 0.64119*x});  
\addplot [>={Latex[length=1.5mm,width=0.5mm,angle'=25,open,round]},,domain=16:18,samples=20,style=semithick,color=green]({-792.0024 + 64.2986*x + -0.64157*x^2},{3614.6146 + -359.331*x + 10.3387*x^2});  
\addplot [>={Latex[length=1.5mm,width=0.5mm,angle'=25,open,round]},,domain=0:1,samples=2,style=semithick,color=green]({157.5058 + 13.4942*x},{496.4103 + 13.5897*x});  
\addplot [>={Latex[length=1.5mm,width=0.5mm,angle'=25,open,round]},,domain=0:1,samples=2,style=semithick,color=green]({171 + 13.3492*x},{510 + -12.5098*x});  
\addplot [>={Latex[length=1.5mm,width=0.5mm,angle'=25,open,round]},,domain=0:1,samples=2,style=semithick,color=green]({184.6387 + 7.3613*x},{496.6222 + -1.6222*x});  
\addplot [>={Latex[length=1.5mm,width=0.5mm,angle'=25,open,round]},,domain=0:1,samples=2,style=semithick,color=green]({192 + 8.0009*x},{495 + -0.64029*x});  
\addplot [>={Latex[length=1.5mm,width=0.5mm,angle'=25,open,round]},,domain=20:21,samples=10,style=semithick,color=green]({-102.5911 + 15.1296*x},{257.6746 + 11.8343*x});  
\addplot [>={Latex[length=1.5mm,width=0.5mm,angle'=25,open,round]},,domain=0:1,samples=2,style=semithick,color=green]({215.1305 + -2.1305*x},{506.194 + -2.194*x});  
\addplot [>={Latex[length=1.5mm,width=0.5mm,angle'=25,open,round]},,domain=0:1,samples=2,style=semithick,color=green]({213 + 16.8755*x},{504 + 2.3263*x});  
\addplot [->,>={Latex[length=1.5mm,width=0.5mm,angle'=25,open,round]},,domain=22:31,samples=90,style=semithick,color=green]({-1787.1528 + 198.9279*x + -6.6387*x^2 + 0.080181*x^3},{1314.1643 + -88.4469*x + 3.2175*x^2 + -0.039374*x^3});  
\end{axis} 
\end{tikzpicture} 

  \noindent}} &
\resizebox {!}{\sz\textwidth} {\resizebox {!}{0.2\textwidth} {\begin{tikzpicture} 
\begin{axis}[y dir=reverse, 
 xmin=1,xmax=960, 
 ymin=1,ymax=600, 
 xticklabels = \empty, yticklabels = \empty, 
 grid=none, axis equal image] 
\addplot graphics[xmin=1,xmax=960,ymin=1,ymax=600] {imgs/thumbnails/tbd/roll_golf.jpg}; 
\addplot [>={Latex[length=1.5mm,width=0.5mm,angle'=25,open,round]},,domain=1:2,samples=10,style=semithick,color=green]({194.4026 + 30.8373*x + -0.93274*x^2},{766.323 + -105.5891*x + 2.3457*x^2});  
\addplot [>={Latex[length=1.5mm,width=0.5mm,angle'=25,open,round]},,domain=2:8,samples=60,style=semithick,color=green]({194.4026 + 30.8373*x + -0.93274*x^2},{766.323 + -105.5891*x + 2.3457*x^2});  
\addplot [>={Latex[length=1.5mm,width=0.5mm,angle'=25,open,round]},,domain=0:1,samples=2,style=semithick,color=green]({381.406 + 2.8045*x},{71.737 + -7.2943*x});  
\addplot [->,>={Latex[length=1.5mm,width=0.5mm,angle'=25,open,round]},,domain=8:17,samples=90,style=semithick,color=green]({196.6121 + 39.8364*x + -2.5041*x^2 + 0.056971*x^3},{-250.6593 + 51.9996*x + -1.81*x^2 + 0.029189*x^3});  
\end{axis} 
\end{tikzpicture} 

  \noindent}} &
\resizebox {!}{\sz\textwidth} {\resizebox {0.33\textwidth}{!} {\begin{tikzpicture} 
\begin{axis}[y dir=reverse, 
 xmin=1,xmax=1920, 
 ymin=1,ymax=1080, 
 xticklabels = \empty, yticklabels = \empty, 
 grid=none, axis equal image] 
\addplot graphics[xmin=1,xmax=1920,ymin=1,ymax=1080] {imgs/thumbnails/tbd/tennis.jpg}; 
 \addplot [>={Latex[length=1.5mm,width=0.5mm,angle'=25,open,round]},,domain=1:3,samples=20,style=semithick,color=green]({1591.5325 + 8.0237*x + -6.3935*x^2 + 0.6162*x^3 + -0.034127*x^4 + 0.00093483*x^5 + -9.7397e-06*x^6},{409.4095 + -9.4851*x + -2.1134*x^2 + 0.29565*x^3 + -0.014904*x^4 + 0.000412*x^5 + -4.7481e-06*x^6});  
\addplot [>={Latex[length=1.5mm,width=0.5mm,angle'=25,open,round]},,domain=3:28,samples=250,style=semithick,color=green]({1591.5325 + 8.0237*x + -6.3935*x^2 + 0.6162*x^3 + -0.034127*x^4 + 0.00093483*x^5 + -9.7397e-06*x^6},{409.4095 + -9.4851*x + -2.1134*x^2 + 0.29565*x^3 + -0.014904*x^4 + 0.000412*x^5 + -4.7481e-06*x^6});  
\addplot [>={Latex[length=1.5mm,width=0.5mm,angle'=25,open,round]},,domain=0:1,samples=2,style=semithick,color=green]({749.4065 + -7.4065*x},{618.4909 + 7.5091*x});  
\addplot [>={Latex[length=1.5mm,width=0.5mm,angle'=25,open,round]},,domain=0:1,samples=2,style=semithick,color=green]({742 + -12.3801*x},{626 + -20.3982*x});  
\addplot [->,>={Latex[length=1.5mm,width=0.5mm,angle'=25,open,round]},,domain=29:39,samples=100,style=semithick,color=green]({2804.6396 + -167.1665*x + 4.7146*x^2 + -0.048881*x^3},{5059.3525 + -348.6253*x + 9.15*x^2 + -0.083595*x^3});  
\end{axis} 
\end{tikzpicture} 

  \noindent}} &
\resizebox {!}{\sz\textwidth} {\resizebox {!}{0.33\textwidth} {\begin{tikzpicture} 
\begin{axis}[y dir=reverse, 
 xmin=1,xmax=960, 
 ymin=1,ymax=600, 
 xticklabels = \empty, yticklabels = \empty, 
 grid=none, axis equal image] 
\addplot graphics[xmin=1,xmax=960,ymin=1,ymax=600] {imgs/thumbnails/tbd/throw_tennis.jpg}; 
\addplot [>={Latex[length=1.5mm,width=0.5mm,angle'=25,open,round]},,domain=1:3,samples=20,style=semithick,color=green]({1089.0183 + -79.2*x + -0.49857*x^2},{67.0235 + 27.1323*x + 7.9453*x^2});  
\addplot [>={Latex[length=1.5mm,width=0.5mm,angle'=25,open,round]},,domain=3:5,samples=20,style=semithick,color=green]({1043.3875 + -54.6304*x + -3.6184*x^2},{108.5527 + 5.0947*x + 10.6768*x^2});  
\addplot [>={Latex[length=1.5mm,width=0.5mm,angle'=25,open,round]},,domain=0:1,samples=2,style=semithick,color=green]({679.7764 + -85.7764*x},{400.9474 + 113.5526*x});  
\addplot [>={Latex[length=1.5mm,width=0.5mm,angle'=25,open,round]},,domain=0:1,samples=2,style=semithick,color=green]({594 + -3.7773*x},{514.5 + 0.63131*x});  
\addplot [>={Latex[length=1.5mm,width=0.5mm,angle'=25,open,round]},,domain=6:13,samples=70,style=semithick,color=green]({1046.0455 + -76.6812*x + 0.11845*x^2},{1428.8778 + -204.5807*x + 8.7149*x^2});  
\addplot [>={Latex[length=1.5mm,width=0.5mm,angle'=25,open,round]},,domain=0:1,samples=2,style=semithick,color=green]({69.2085 + -12.2085*x},{242.1527 + 5.3473*x});  
\addplot [>={Latex[length=1.5mm,width=0.5mm,angle'=25,open,round]},,domain=0:1,samples=2,style=semithick,color=green]({57 + 7.305*x},{247.5 + 8.1383*x});  
\addplot [>={Latex[length=1.5mm,width=0.5mm,angle'=25,open,round]},,domain=14:18,samples=40,style=semithick,color=green]({-559.6904 + 35.6583*x + 0.63663*x^2},{1780.3085 + -225.7891*x + 8.3489*x^2});  
\addplot [>={Latex[length=1.5mm,width=0.5mm,angle'=25,open,round]},,domain=0:1,samples=2,style=semithick,color=green]({288.4268 + 64.0732*x},{421.1362 + 87.3638*x});  
\addplot [>={Latex[length=1.5mm,width=0.5mm,angle'=25,open,round]},,domain=0:1,samples=2,style=semithick,color=green]({352.5 + 2.245*x},{508.5 + 10.0542*x});  
\addplot [>={Latex[length=1.5mm,width=0.5mm,angle'=25,open,round]},,domain=19:27,samples=80,style=semithick,color=green]({-29.2702 + 15.1045*x + 0.26878*x^2},{4765.3426 + -379.5912*x + 8.2145*x^2});  
\addplot [>={Latex[length=1.5mm,width=0.5mm,angle'=25,open,round]},,domain=0:1,samples=2,style=semithick,color=green]({574.4917 + 9.0083*x},{504.7703 + 17.2297*x});  
\addplot [>={Latex[length=1.5mm,width=0.5mm,angle'=25,open,round]},,domain=0:1,samples=2,style=semithick,color=green]({583.5 + 18.7376*x},{522 + -26.0871*x});  
\addplot [>={Latex[length=1.5mm,width=0.5mm,angle'=25,open,round]},,domain=28:33,samples=50,style=semithick,color=green]({-343.3295 + 38.5068*x + -0.16916*x^2},{8489.8874 + -526.4252*x + 8.6045*x^2});  
\addplot [>={Latex[length=1.5mm,width=0.5mm,angle'=25,open,round]},,domain=0:1,samples=2,style=semithick,color=green]({743.1772 + 21.8228*x},{488.1604 + 38.3396*x});  
\addplot [>={Latex[length=1.5mm,width=0.5mm,angle'=25,open,round]},,domain=0:1,samples=2,style=semithick,color=green]({765 + 3.8735*x},{526.5 + -3.4273*x});  
\addplot [>={Latex[length=1.5mm,width=0.5mm,angle'=25,open,round]},,domain=34:39,samples=50,style=semithick,color=green]({-2139.1446 + 138.3709*x + -1.5541*x^2},{10004.3845 + -519.8301*x + 7.0873*x^2});  
\addplot [>={Latex[length=1.5mm,width=0.5mm,angle'=25,open,round]},,domain=0:1,samples=2,style=semithick,color=green]({893.4649 + -19.4431*x},{510.7842 + 18.4587*x});  
\addplot [>={Latex[length=1.5mm,width=0.5mm,angle'=25,open,round]},,domain=39:41,samples=20,style=semithick,color=green]({-5546.532 + 331.0936*x + -4.2683*x^2},{11806.4449 + -554.7971*x + 6.8112*x^2});  
\addplot [>={Latex[length=1.5mm,width=0.5mm,angle'=25,open,round]},,domain=0:1,samples=2,style=semithick,color=green]({853.2797 + -17.7797*x},{509.4458 + 17.0542*x});  
\addplot [>={Latex[length=1.5mm,width=0.5mm,angle'=25,open,round]},,domain=0:1,samples=2,style=semithick,color=green]({835.5 + 2.5922*x},{526.5 + 1.8661*x});  
\addplot [>={Latex[length=1.5mm,width=0.5mm,angle'=25,open,round]},,domain=42:43,samples=10,style=semithick,color=green]({1492.5246 + -15.5817*x},{1019.1907 + -11.6863*x});  
\addplot [>={Latex[length=1.5mm,width=0.5mm,angle'=25,open,round]},,domain=0:1,samples=2,style=semithick,color=green]({822.5105 + -0.5105*x},{516.6798 + -0.67979*x});  
\addplot [>={Latex[length=1.5mm,width=0.5mm,angle'=25,open,round]},,domain=0:1,samples=2,style=semithick,color=green]({822 + -16.7429*x},{516 + 13.9434*x});  
\addplot [>={Latex[length=1.5mm,width=0.5mm,angle'=25,open,round]},,domain=0:1,samples=2,style=semithick,color=green]({805.1919 + -14.6919*x},{526.014 + -4.014*x});  
\addplot [>={Latex[length=1.5mm,width=0.5mm,angle'=25,open,round]},,domain=0:1,samples=2,style=semithick,color=green]({790.5 + -0.15606*x},{522 + -0.29684*x});  
\addplot [->,>={Latex[length=1.5mm,width=0.5mm,angle'=25,open,round]},,domain=45:46,samples=10,style=semithick,color=green]({1483.275 + -15.3985*x},{157.4216 + 8.0951*x});  
\end{axis} 
\end{tikzpicture} 

  \noindent}} \\

\rotatebox{90}{\hspace{6pt} Ground Truth} &
\resizebox {!}{\sz\textwidth} {\input{imgs/gt/hit_tennis}} &
\resizebox {!}{\sz\textwidth} {\input{imgs/gt/roll_golf}} &
\resizebox {!}{\sz\textwidth} {\input{imgs/gt/tennis}} &
\resizebox {!}{\sz\textwidth} {\input{imgs/gt/throw_tennis}} \\

\end{tabular}
\caption{Trajectory estimation on four sequences from the TbD dataset. From top to bottom: real-time version of the proposed method, with deblurring, with deblurring and non-causal post-processing from~\cite{tbdnc}, TbD~\cite{tbd}, TbD-NC~\cite{tbdnc}, the ground truth. }
\label{fig:res_tbd}
\end{figure*}

\subsection{Applications}
The main application is FMO retrieval as shown on YouTube videos in Fig.~\ref{fig:retrieval}. 
Processing both videos (total 15 minutes) took us 15 minutes, whereas the TbD method took around 25 hours.
Another application is temporal super-resolution where we generate videos with higher frame rates.
After $F, M, H, \mathcal{C}(t)$ are extracted, we split the fitted trajectory $\mathcal{C}(t)$ into the desired number of pieces and apply the formation model~\eqref{eq:model} to each temporal super-resolution frame separately with the same $F,M$.
Examples are shown in the supplementary material.

\subsection{Model generalization}
As shown in previous experiments, the method works well on real data, even when trained on synthetic. 
The method has only seen synthetic FMOs with spherical shape in the training examples, but FMOs of more complex shapes can also be detected. 
Examples are in Fig.~\ref{fig:retrieval} (a hand, a cap, and a keychain), in Fig.~\ref{fig:res_deblur} (thrown wooden cube), and in Fig.~\ref{fig:falling} (marker). 
Since FMODetect is trained with dynamic cameras and backgrounds from the VOT dataset, it is robust to the camera and background motion (Fig.~\ref{fig:retrieval}).
All previous methods fail to detect or reconstruct objects with low contrast (softball in Fig.~\ref{fig:res_deblur}), while the proposed method succeeds in this task as there was no restriction on the contrast in the synthetic dataset.
Among drawbacks of the proposed method are failures on irregular motions,~\eg the ceiling fan in Fig.~\ref{fig:retrieval} since the synthesized dataset has no such motions. 
FMO deblurring from a single input image without the background is an open problem~--~the need for the background image is currently a limitation, but it can be computed for many applications.

\section{Conclusion}
We proposed the first learning-based method for fast moving object detection. 
FMODetect improves precision, robustness, and computation time compared to previous methods. 
In terms of detection accuracy, the proposed methods achieves significantly better results on challenging datasets.
By simplifying deblatting into matting and deblurring, we estimate appearance using energy minimization and produce qualitatively better results than deblatting. The proposed method is real-time capable and an order of magnitude faster than previous FMO detection methods. This allows for realistic applications such as efficient FMO scanning or retrieval in large video collections. 
\newcommand\blfootnote[1]{%
  \begingroup
  \renewcommand\thefootnote{}\footnote{#1}%
  \addtocounter{footnote}{-1}%
  \endgroup
}
\blfootnote{\textbf{Acknowledgements.} This research was supported by Research Center for Informatics (project CZ$.02.1.01/0.0/0.0/16\_019/0000765$ funded by OP VVV), Google Focused Research Award, the Czech Science Foundation grant GA21-03921S, the Praemium Academiae awarded
by the Czech Academy of Sciences, International Federation of Association Football (FIFA), and Innosuisse grant No. 34475.1 IP-ICT.}

\newpage

{\small
\bibliographystyle{ieee_fullname}
\bibliography{egbib}
}

\end{document}